\documentclass[10pt,twocolumn,letterpaper]{article}

\usepackage{cvpr}              

\usepackage{soul}

\definecolor{cvprblue}{rgb}{0.21,0.49,0.74}
\usepackage[pagebackref,breaklinks,colorlinks,allcolors=cvprblue]{hyperref}
\usepackage[most]{tcolorbox}

\def\paperID{10876} 
\def\confName{CVPR}
\def\confYear{2025}

\usepackage{hyperref}
\usepackage{url}

\usepackage{colortbl}
\usepackage{xcolor}
\usepackage{color}
\usepackage{xspace}
\usepackage{pifont}
\usepackage{arydshln}
\usepackage[T1]{fontenc}
\definecolor{citecolor}{HTML}{0071BC}
\definecolor{linkcolor}{HTML}{ED1C24}
\definecolor{LGray}{gray}{0.97}
\usepackage{wrapfig}
\usepackage{float}
\usepackage{xtab}

\usepackage[capitalize]{cleveref}

\definecolor{darkgreen}{rgb}{0.0, 0.5, 0.0}  

\newcommand{\ALMbench}{\texttt{ALM-bench}\xspace}

\title{All Languages Matter: Evaluating LMMs on Culturally Diverse 100 Languages }

\author{
    \begin{minipage}[t]{\textwidth}
        \centering
         {Ashmal Vayani}\textsuperscript{1, 2 \thanks{ashmal.vayani@ucf.edu}}, 
         {Dinura Dissanayake}\textsuperscript{2}, 
         {Hasindri Watawana}\textsuperscript{2}, 
         {Noor Ahsan}\textsuperscript{2},
         {Nevasini Sasikumar}\textsuperscript{2}, 
         {Omkar Thawakar}\textsuperscript{2}, 
         {Henok Biadglign Ademtew}\textsuperscript{7}, 
         {Yahya Hmaiti}\textsuperscript{1},
         {Amandeep Kumar}\textsuperscript{8}, 
         {Kartik Kuckreja}\textsuperscript{2},
         {Mykola Maslych}\textsuperscript{1}, 
         {Wafa Al Ghallabi}\textsuperscript{2}, 
         {Mihail Mihaylov}\textsuperscript{2},
         {Chao Qin}\textsuperscript{2},
         {Abdelrahman M Shaker}\textsuperscript{2}, 
         {Mike Zhang}\textsuperscript{9}, 
         {Mahardika Krisna Ihsani}\textsuperscript{2}, 
         {Amiel Esplana}\textsuperscript{10}, 
         {Monil Gokani}\textsuperscript{11}, 
         {Shachar Mirkin}\textsuperscript{12},
         {Harsh Singh}\textsuperscript{2}, 
         {Ashay Srivastava}\textsuperscript{13}, 
         {Endre Hamerlik}\textsuperscript{14},
         {Fathinah Asma Izzati}\textsuperscript{2}, 
         {Fadillah Adamsyah Maani}\textsuperscript{2}, 
         {Sebastian Cavada}\textsuperscript{2}, 
         {Jenny Chim}\textsuperscript{15}, 
         {Rohit Gupta}\textsuperscript{1},
         {Sanjay Manjunath}\textsuperscript{2}, 
         {Kamila Zhumakhanova}\textsuperscript{2},
         {Feno Heriniaina Rabevohitra}\textsuperscript{16}, 
         {Azril Amirudin}\textsuperscript{17},
         {Muhammad Ridzuan}\textsuperscript{2}, 
         {Daniya Kareem}\textsuperscript{2}, 
         {Ketan More}\textsuperscript{2}, 
         {Kunyang Li}\textsuperscript{1}, 
         {Pramesh Shakya}\textsuperscript{1},
         {Muhammad Saad}\textsuperscript{2}, 
         {Amirpouya Ghasemaghaei}\textsuperscript{1}, 
         {Amirbek Djanibekov}\textsuperscript{2},
         {Dilshod Azizov}\textsuperscript{2},
         {Branislava Jankovic}\textsuperscript{2},
         {Naman Bhatia}\textsuperscript{11}, 
         {Alvaro Cabrera}\textsuperscript{2}, 
         {Johan Obando-Ceron}\textsuperscript{18}, 
         {Olympiah Otieno}\textsuperscript{19},
         {Fabian Farestam}\textsuperscript{20},  
         {Muztoba Rabbani}\textsuperscript{21}, 
         {Sanoojan Baliah}\textsuperscript{2}, 
         {Santosh Sanjeev}\textsuperscript{2},
         {Abduragim Shtanchaev}\textsuperscript{2},
         {Maheen Fatima}\textsuperscript{22}, 
         {Thao Nguyen}\textsuperscript{2}, 
         {Amrin Kareem}\textsuperscript{2}, 
         {Toluwani Aremu}\textsuperscript{2}, 
         {Nathan Xavier}\textsuperscript{23},
         {Amit Bhatkal}\textsuperscript{2},
         {Hawau Toyin}\textsuperscript{2},
         {Aman Chadha}\textsuperscript{3}, 
         {Hisham Cholakkal}\textsuperscript{2 },
         {Rao Muhammad Anwer}\textsuperscript{2, 4 },
         {Michael Felsberg}\textsuperscript{6 }, 
         {Jorma Laaksonen}\textsuperscript{4 },
         {Thamar Solorio}\textsuperscript{2 },
         {Monojit Choudhury}\textsuperscript{2 },
         {Ivan Laptev}\textsuperscript{2},
         {Mubarak Shah}\textsuperscript{1},
         {Salman Khan}\textsuperscript{2, 5}, 
         {Fahad Shahbaz Khan}\textsuperscript{2, 6} \\ [1em]
        \normalsize{
            \textsuperscript{1}University of Central Florida, 
            \textsuperscript{2}Mohamed bin Zayed University of AI, 
            \textsuperscript{3}Amazon, 
            \textsuperscript{4}Aalto University, 
            \textsuperscript{5}Australian National University, 
            \textsuperscript{6}Linköping University,
            \textsuperscript{7}Ehtiopian AI Institute,
            \textsuperscript{8}Johns Hopkins University,
            \textsuperscript{9}Aalborg University,
            \textsuperscript{10}University of the West of England,
            \textsuperscript{11}IIT Hyderadabad,
            \textsuperscript{12}Alpinference,
            \textsuperscript{13}University of Maryland,
            \textsuperscript{14}HUN-REN Institute for CS and Control,
            \textsuperscript{15}Queen Mary University of London,
            \textsuperscript{16}Chongqing University,
            \textsuperscript{17}University of the People,
            \textsuperscript{18}University of Montreal,
            \textsuperscript{19}JKUAT,
            \textsuperscript{20}ETH Zurich,
            \textsuperscript{21}University of California, Merced,
            \textsuperscript{22}Air University,
            \textsuperscript{23}Universidade Federal de Minas Gerais
            }
        \end{minipage}
    \vspace{-1em}
}

\begin{document}

\twocolumn[{
\renewcommand\twocolumn[1][]{#1}
\maketitle
\begin{center}
\vspace{-0.4em}
\captionsetup{type=figure}
\includegraphics[width=\textwidth]{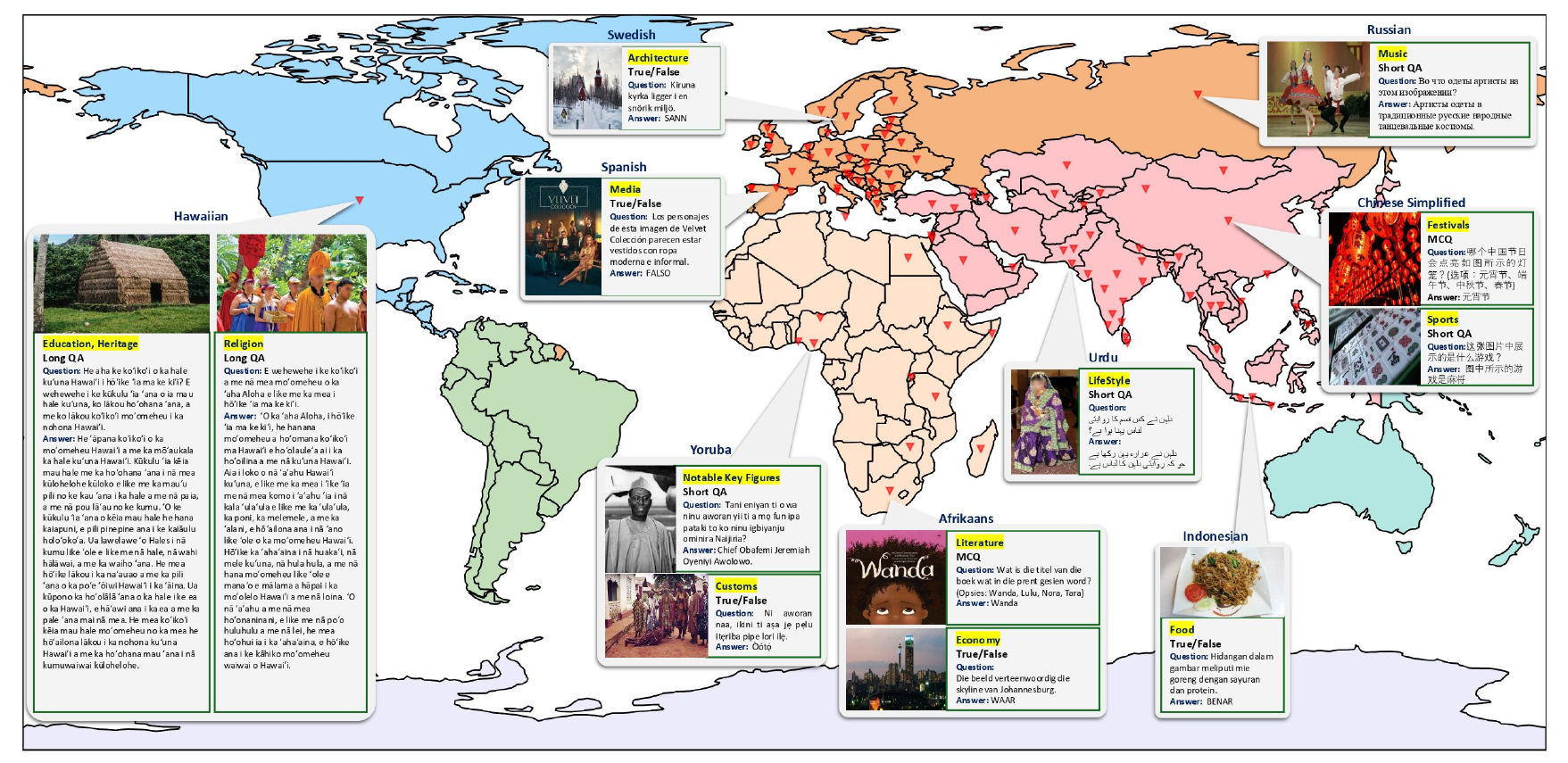}\vspace{-1em}
\captionof{figure}{\ALMbench comprises a diverse set of 100 languages with manually verified annotations by respective native language experts. Here, qualitative examples highlight the comprehensive set of 13 cultural aspects covered in the benchmark, such as heritage, customs, architecture, literature, music, and sports. 
It also evaluates visual understanding for six generic aspects. 
The \ALMbench focuses on low-resource languages and different regions, spanning 73 countries across five continents and 24 distinct scripts. \ALMbench covers diverse questions, such as multiple choice questions (MCQs), true/false (T/F), short and long visual question answers (VQAs). 
}
\label{fig_Intro}
\vspace{-0.5em}
\end{center}
}]

\begin{figure*}[t!]
    \centering
    \begin{subfigure}[b]{0.54\textwidth}
        \centering
        \includegraphics[width=\textwidth]{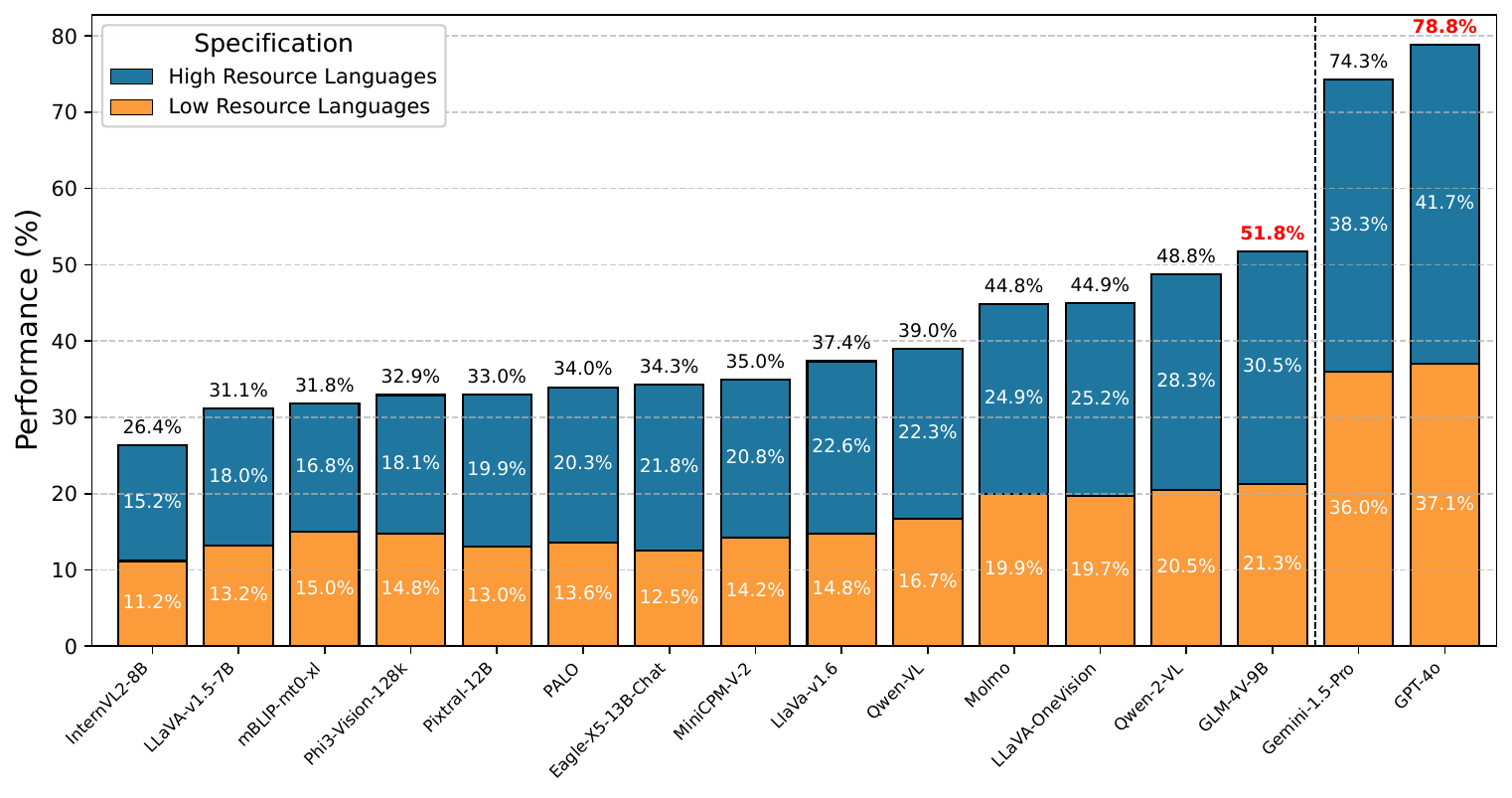}
        \caption{Performance comparison of open vs. closed-source models.}
        \label{fig:open_vs_close}
    \end{subfigure}
    \hfill
    \begin{subfigure}[b]{0.40\textwidth}
        \centering
        \includegraphics[width=0.8\textwidth]{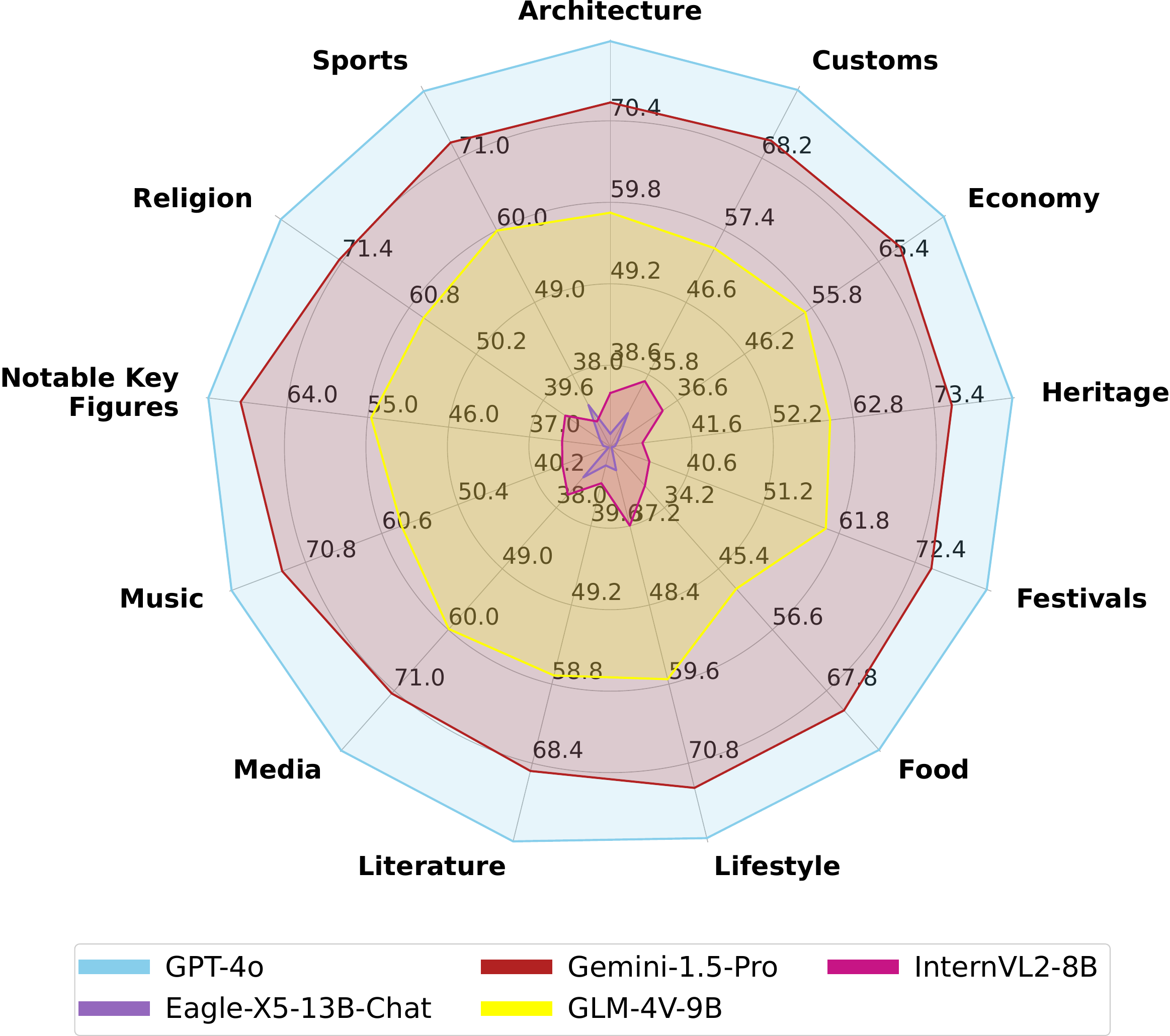}
        \caption{Performance on various cultural categories.}
        \label{fig:category_wise}
    \end{subfigure}
    \caption{Benchmarking LMMs on diverse languages \& cultures: (a) Performance of various open versus closed-sourced LMMs on \ALMbench. For each LMM, we also report performance on low versus high-resource languages. All these carefully selected models were released after 2023. (b) Performance of high-performing LMMs on our culturally curated 13 categories available in our \ALMbench.}
    \label{fig:category_wise_performance}
    
\end{figure*}

 \begin{abstract}
\footnotetext[1]{corresponding author: ashmal.vayani@ucf.edu} 
Existing Large Multimodal Models (LMMs) generally focus on only a few regions and languages. 
As LMMs continue to improve, it is increasingly important to ensure they understand cultural contexts, respect local sensitivities, and support low-resource languages, all while effectively integrating corresponding visual cues.
In pursuit of culturally diverse global multimodal models, our proposed All Languages Matter Benchmark (\ALMbench) represents the largest and most comprehensive effort to date for evaluating LMMs across 100 languages. 
\ALMbench challenges existing models by testing their ability to understand and reason about culturally diverse images paired with text in various languages, including many low-resource languages traditionally underrepresented in LMM research. 
The benchmark offers a robust and nuanced evaluation framework featuring various question formats, including true/false, multiple choice, and open-ended questions, which are further divided into short and long-answer categories. 
\ALMbench design ensures a comprehensive assessment of a model’s ability to handle varied levels of difficulty in visual and linguistic reasoning. 
To capture the rich tapestry of global cultures, \ALMbench carefully curates content from 13 distinct cultural aspects, ranging from traditions and rituals to famous personalities and celebrations. 
Through this, \ALMbench not only provides a rigorous testing ground for state-of-the-art open and closed-source LMMs but also highlights the importance of cultural and linguistic inclusivity, encouraging the development of models that can serve diverse global populations effectively. Our benchmark is publicly available at \url{https://mbzuai-oryx.github.io/ALM-Bench/}.
\end{abstract}    
\section{Introduction}
\label{sec:intro}

In recent years, large multimodal models (LMMs) have seen significant progress across a variety of vision-language tasks \cite{achiam2023gpt, team2023gemini, Qwen-VL, chen2024internvl, Maaz2023VideoChatGPT, hanoona2023GLaMM}. 
Despite these advances, a major limitation remains in the models’ capacity to understand and respond accurately to cultural and linguistic diversity \cite{romero2024cvqa}. 
Current LMMs perform well in widely spoken languages with substantial training data but struggle with low-resource languages and nuanced cultural contexts, limiting their effectiveness on a global scale \cite{maaz2024palo}.

Existing benchmarks for evaluating multilingual and multicultural capabilities in LMMs are limited in scope, often representing only a small subset of cultures and languages.
Such benchmarks generally focus on high-resource languages and overlook the cultural richness of low-resource languages, which hinders the development and evaluation of models that can serve diverse populations across the globe.
For instance, CulturalVQA \cite{nayak2024benchmarking} and GD-VCR \cite{yin2021broaden} focus on culturally specific content in English and do not reflect global cultural diversity.
Similarly, the Henna benchmark \cite{alwajih2024peacock} includes cultural information from 11 countries but is restricted to the Arabic language. 
In terms of multilingualism, existing benchmarks \cite{sun2024parrot, liu2023mmbench, das2024exams, zhang2023m3exam, wang2024m4u, fu2024mmecomprehensiveevaluationbenchmark} include multilingual data but remain constrained by limited cultural diversity and are often restricted to only single question type.
This gap highlights the need for a more comprehensive benchmark that addresses both linguistic and cultural inclusivity in LMMs.

To address these challenges, we introduce a culturally and linguistically diverse multimodal benchmark: All Languages Matter Benchmark (\ALMbench). 
\ALMbench is specifically designed to evaluate LMMs across a diverse set of 100 languages, including many low-resource ones, with a strong emphasis on cultural depth. The complete list of these languages is presented in the Tab. A. 3. (suppl. material). In comparison to the recent CVQA benchmark \cite{romero2024cvqa}, \ALMbench offers 3x more languages, more diverse question types, and $\sim$2x more image domains and samples with an extensive focus on fine-grained cultural aspects (Fig.~\ref{fig_Intro}).
With these enhancements, \ALMbench aims to assess the next generation of massively multilingual multimodal models in a standardized way, pushing the boundaries of LMMs towards better cultural understanding and inclusivity.

Our evaluations with 16 recent state-of-the-art LMMs show that existing models suffer from low-resource languages and cultures (Fig.~\ref{fig:category_wise_performance}). 
The analysis also shows a significant gap between the open and closed-source models in understanding cultural and linguistic nuances in a multimodal setting; e.g., GPT4o performs 27\%  better than the best open-source model (GLM-4V). 
Our evaluations across scripts and language groups highlight the need to improve the understanding of underrepresented regions, such as Southeast Asia and West Africa. 
We also show how simple prompting techniques can help improve LMM performance.

Our contributions can be summarized as follows:
\begin{itemize}[itemsep=2pt, parsep=1pt, topsep=0pt]
    \item We introduce \ALMbench, a culturally diverse multilingual and multimodal VQA benchmark covering 100 languages with 22.7K question-answers. \ALMbench encompasses 19 generic and culture-specific domains for each language, enriched with four diverse question types.
    \item \ALMbench is meticulously curated and verified with native-language experts (over 800 hours of human
annotators), ensuring cultural relevance and accuracy across low- and high-resource languages alike.
    \item We benchmark existing LMMs on \ALMbench, identifying performance gaps and areas for improvement, especially in culturally complex multilingual scenarios. 
\end{itemize}



\section{Related Work}
\label{sec:relatedwork}

To align large language models (LLMs) with human preferences, various approaches have been developed \cite{ziegler2019fine,rafailov2024direct,schulman2017proximal}. Over time, these methods are extended to enable LLMs to be sensitive to geo-diverse cultural values \cite{abbasi2023persianllama, lin2023taiwan, narnaware2025sb, sirnam2024x}. 
This evolution has broadened their applicability to both multilingual and multimodal LLMs, enabling greater engagement with users from diverse linguistic and cultural backgrounds.
Despite this progress, LMM benchmarks often fail to adequately address cultural aspects, especially for low-resource languages.
Models such as GPT-4V \cite{achiam2023gpt} and
 Gemini 1.5 Pro \cite{team2023gemini} performs well on existing multimodal benchmarks, but these evaluations often lack in capturing the intricacies of multilingual and culturally diverse interactions.
Open-source models, in particular, face challenges in accurately representing geodiverse  values (e.g., local beliefs, rituals, and traditions), underscoring the need for a culturally aware evaluation framework for LMMs (see Tab. \ref{tab:methods_comparison}).

Several recent benchmarks have been developed to evaluate the cultural aspects of multilingual LMMs, such as, the Henna benchmark \cite{alwajih2024peacock} representing cultural information from 11 countries with nearly 1,132 samples. However, it is restricted to the Arabic language, covering only five domains while being closed-source. Other benchmarks, such as CulturalVQA \cite{nayak2024benchmarking} and GD-VCR \cite{yin2021broaden} are also small-scale and focus only on culturally specific content. These benchmarks are limited to English and struggle to encompass the full spectrum of global cultural diversity. Their samples are unevenly distributed across countries and domains, which may lead to skewed assessment.

Benchmarks such as SEED-Bench \cite{li2024seed, li2023seed} and MMStar \cite{chen2024we} extend the CC3M dataset \cite{ng2020understanding} to cover an extended set of domains. 
Despite their broader scope, these benchmarks lack cultural nuance and are constrained to multiple-choice questions (MCQ) which only assess the LMMs in one dimension. Although MM-Vet \cite{yu2023mm} and MMMU \cite{yue2024mmmu} offer a variety of question types, they primarily address basic perceptual abilities without demanding in-depth cultural reasoning or expert-level domain knowledge.

For multilingual LMM benchmarking, CVQA \cite{romero2024cvqa} represents a recent effort, collecting 
cultural samples across 31 languages. However, it is limited to a single question type, MCQ, and features only a small proportion of samples on traditions and rituals. 
Similarly, MaRVL \cite{liu2021visually} covers five languages but lacks probing into cultural common sense and primarily focuses on non-Western cultural information, resulting in limited diversity. Other benchmarks, including MMMB \cite{sun2024parrot}, MMBench \cite{liu2023mmbench}, Exams-V \cite{das2024exams}, M3Exam \cite{zhang2023m3exam}, M4U \cite{wang2024m4u}, and MME \cite{fu2024mmecomprehensiveevaluationbenchmark}, do include multilingual evaluation samples. However, they are constrained by a lack of cultural diversity in their datasets, cover fewer languages, and are restricted to a single question type (e.g., MCQs).

Our proposed benchmark, \ALMbench, bridges the aforementioned gaps (see Tab. \ref{tab:methods_comparison}) by introducing a culturally diverse dataset covering Western and non-Western regions, low and high-resource languages (100 in total), and specific regional dialects (e.g., Emirati Arabic, Egyptian Arabic). \ALMbench provides a robust evaluation framework for Multilingual LMMs, enabling a comprehensive assessment of their generic and cultural adaptability and promoting inclusivity in multilingual multimodal AI research.

\begin{table*} 

\centering
    \setlength{\tabcolsep}{4pt}
    \resizebox{\textwidth}{!}{%
    \begin{tabular}{lcccccccccccccc}
        \toprule
        \textbf{Benchmark} & \textbf{Multilingual} & \textbf{\# of} & \textbf{\# of} & \textbf{Total} & \textbf{Total}  & \textbf{Question} & \textbf{Question} & \textbf{Annotation} & \textbf{Cultural Specific} & \textbf{Open-} & \textbf{Bias} & \textbf{Diversity} \\
        & & \textbf{Languages} & \textbf{Scripts} & \textbf{Domains} & \textbf{Samples}  & \textbf{Types} & \textbf{Forms} & \textbf{Type} & \textbf{Content} & \textbf{Source} & \textbf{Correction}
        \\
        \midrule

        \rowcolor{LGray} Henna \cite{alwajih2024peacock} & \textcolor{red}{\ding{55}} & 1 & 1 & 5 & 1132  & SVQA, LVQA & - & Auto+Manual & \textcolor{darkgreen}{\checkmark} & \textcolor{red}{\ding{55}} & \textcolor{darkgreen}{\checkmark} & \textcolor{red}{\ding{55}} \\

        CulturalVQA \cite{nayak2024benchmarking} & \textcolor{red}{\ding{55}} & 1 & 1 & 5 & 2378  & SVQA & - & Manual & \textcolor{darkgreen}{\checkmark} & \textcolor{red}{\ding{55}} & - & \textcolor{red}{\ding{55}} \\

        \rowcolor{LGray} GD-VCR \cite{yin2021broaden} & \textcolor{red}{\ding{55}} & 1 & 1 & 1 & 886  & MCQ & Fixed & Manual & \textcolor{darkgreen}{\checkmark} & \textcolor{darkgreen}{\checkmark} & \textcolor{red}{\ding{55}} & \textcolor{darkgreen}{\checkmark} \\

        SEED-Bench \cite{li2023seed} & \textcolor{red}{\ding{55}} & 1 & 1 & 12 & 19242  & MCQ & Diverse & Auto+Manual & \textcolor{red}{\ding{55}} & \textcolor{darkgreen}{\checkmark} & \textcolor{red}{\ding{55}} & \textcolor{red}{\ding{55}} \\

        \rowcolor{LGray} SEED-Bench2 \cite{li2024seed} & \textcolor{red}{\ding{55}} & 1 & 1 & 34 & 24371  & MCQ & Diverse & Auto+Manual & \textcolor{red}{\ding{55}} & \textcolor{darkgreen}{\checkmark} & \textcolor{red}{\ding{55}} & \textcolor{darkgreen}{\checkmark} \\

        MMStar \cite{chen2024we} & \textcolor{red}{\ding{55}} & 1 & 1 & 18 & 1500  & MCQ & Fixed & Auto+Manual & \textcolor{red}{\ding{55}} & \textcolor{darkgreen}{\checkmark} & \textcolor{darkgreen}{\checkmark} & \textcolor{red}{\ding{55}} \\

        \rowcolor{LGray} MM-Vet \cite{yu2023mm} & \textcolor{red}{\ding{55}} & 1 & 1 & 16 & 218  & SVQA,LVQA & Fixed & Manual & \textcolor{red}{\ding{55}} & \textcolor{darkgreen}{\checkmark} & \textcolor{red}{\ding{55}} & \textcolor{red}{\ding{55}} \\

        MMMU \cite{yue2024mmmu} & \textcolor{red}{\ding{55}} & 1 & 1 & 30 & 11550  & MCQ,SVQA & Fixed & Manual & \textcolor{red}{\ding{55}} & \textcolor{darkgreen}{\checkmark} & \textcolor{darkgreen}{\checkmark} & \textcolor{darkgreen}{\checkmark} \\

        \arrayrulecolor{black}
        \cdashline{1-13}[2pt/2.5pt]

        \rowcolor{LGray} CVQA \cite{romero2024cvqa} & \textcolor{darkgreen}{\checkmark} & 31 & 13 & 10 & 10000  & MCQ & Fixed & Manual & \textcolor{darkgreen}{\checkmark} & \textcolor{darkgreen}{\checkmark} & \textcolor{red}{\ding{55}} & \textcolor{darkgreen}{\checkmark} \\

        MMMB \cite{sun2024parrot} & \textcolor{darkgreen}{\checkmark} & 6 & 4 & 15 & 12000  & TF, MCQ & Fixed & Auto+Manual & \textcolor{red}{\ding{55}} & \textcolor{darkgreen}{\checkmark} & \textcolor{darkgreen}{\checkmark} & \textcolor{darkgreen}{\checkmark} \\

        \rowcolor{LGray} MMBench \cite{liu2023mmbench} & \textcolor{darkgreen}{\checkmark} & 2 & 2 & 20 & 2974  & MCQ & Fixed & Auto+Manual & \textcolor{red}{\ding{55}} & \textcolor{darkgreen}{\checkmark} & \textcolor{darkgreen}{\checkmark} & \textcolor{red}{\ding{55}} \\
        
        EXAMS-V \cite{das2024exams} & \textcolor{darkgreen}{\checkmark} & 11 & 4 & 20 & 20932  & MCQ & Diverse & Manual & \textcolor{red}{\ding{55}} & \textcolor{darkgreen}{\checkmark} & \textcolor{red}{\ding{55}} & \textcolor{darkgreen}{\checkmark} \\

        \rowcolor{LGray} MaRVL \cite{liu2021visually} & \textcolor{darkgreen}{\checkmark} & 5 & 3 & 11 & 7630  & TF & Fixed & Manual & \textcolor{darkgreen}{\checkmark} & \textcolor{darkgreen}{\checkmark} & \textcolor{darkgreen}{\checkmark} & \textcolor{red}{\ding{55}} \\

        M3Exam \cite{zhang2023m3exam} & \textcolor{darkgreen}{\checkmark} & 9 & 3 & 4 & 12317  & MCQ & Diverse & Auto+Manual & \textcolor{red}{\ding{55}} & \textcolor{darkgreen}{\checkmark} & \textcolor{red}{\ding{55}} & \textcolor{darkgreen}{\checkmark} \\

        \rowcolor{LGray} MaXM \cite{changpinyo2022maxm} & \textcolor{darkgreen}{\checkmark} & 7 & 5 & - & 2142  & SVQA & Fixed & Auto+Manual & \textcolor{red}{\ding{55}} & \textcolor{darkgreen}{\checkmark} & \textcolor{darkgreen}{\checkmark} & \textcolor{darkgreen}{\checkmark} \\

        xGQA \cite{pfeiffer2021xgqa} & \textcolor{darkgreen}{\checkmark} & 8 & 5 & - & 12578  & Y/N,SVQA & Fixed & Manual & \textcolor{red}{\ding{55}} & \textcolor{darkgreen}{\checkmark} & \textcolor{red}{\ding{55}} & \textcolor{darkgreen}{\checkmark}  \\

        \rowcolor{LGray} M4U \cite{wang2024m4u} & \textcolor{darkgreen}{\checkmark} & 3 & 2 & 64 & 8931  & MCQ & Fixed & Auto+Manual & \textcolor{red}{\ding{55}} & \textcolor{darkgreen}{\checkmark} & \textcolor{darkgreen}{\checkmark} & \textcolor{red}{\ding{55}}  \\

        MME \cite{fu2024mmecomprehensiveevaluationbenchmark} & \textcolor{darkgreen}{\checkmark} & 2 & 2 & 14 & 2370  & Y/N & Fixed & Manual & \textcolor{red}{\ding{55}} & \textcolor{darkgreen}{\checkmark} & \textcolor{red}{\ding{55}} & \textcolor{red}{\ding{55}}  \\

        


        \rowcolor{violet!10}\textbf{Ours} & \textcolor{darkgreen}{\checkmark} & 100 & 24 & 19 & 22763  & MCQ,SVQA,LVQA,TF & Diverse & Auto+Manual & \textcolor{darkgreen}{\checkmark} & \textcolor{darkgreen}{\checkmark} & \textcolor{darkgreen}{\checkmark} & \textcolor{darkgreen}{\checkmark} \\
        
        
        \bottomrule
    \end{tabular}
    }
    \caption{Comparison of various LMM benchmarks with a focus on multilingual and cultural understanding. The \textit{Domains} indicate the range of aspects covered by the dataset for each language. \textit{Question Form} is categorized as "Diverse" if the questions phrasing varies, and "Fixed" otherwise. \textit{Annotation Types} are classified as "Manual" if questions were originally in the local language, "Manual+Auto" if questions were generated or translated using GPT-4/Google API and subsequently validated by human experts, and "Auto" if generated or translated automatically without human validation. \textit{Bias Correction} reflects whether the dataset is balanced across cultures and countries, while \textit{Diversity} indicates whether the dataset includes both Western and non-Western minority cultures. `-' means information not available.}
    \label{tab:methods_comparison}
\end{table*}

\label{sec:dataset_collection}

\section{ALM Benchmark}

\begin{figure}[!htp]
    \centering
    \includegraphics[scale=0.3, width=0.35\textwidth]{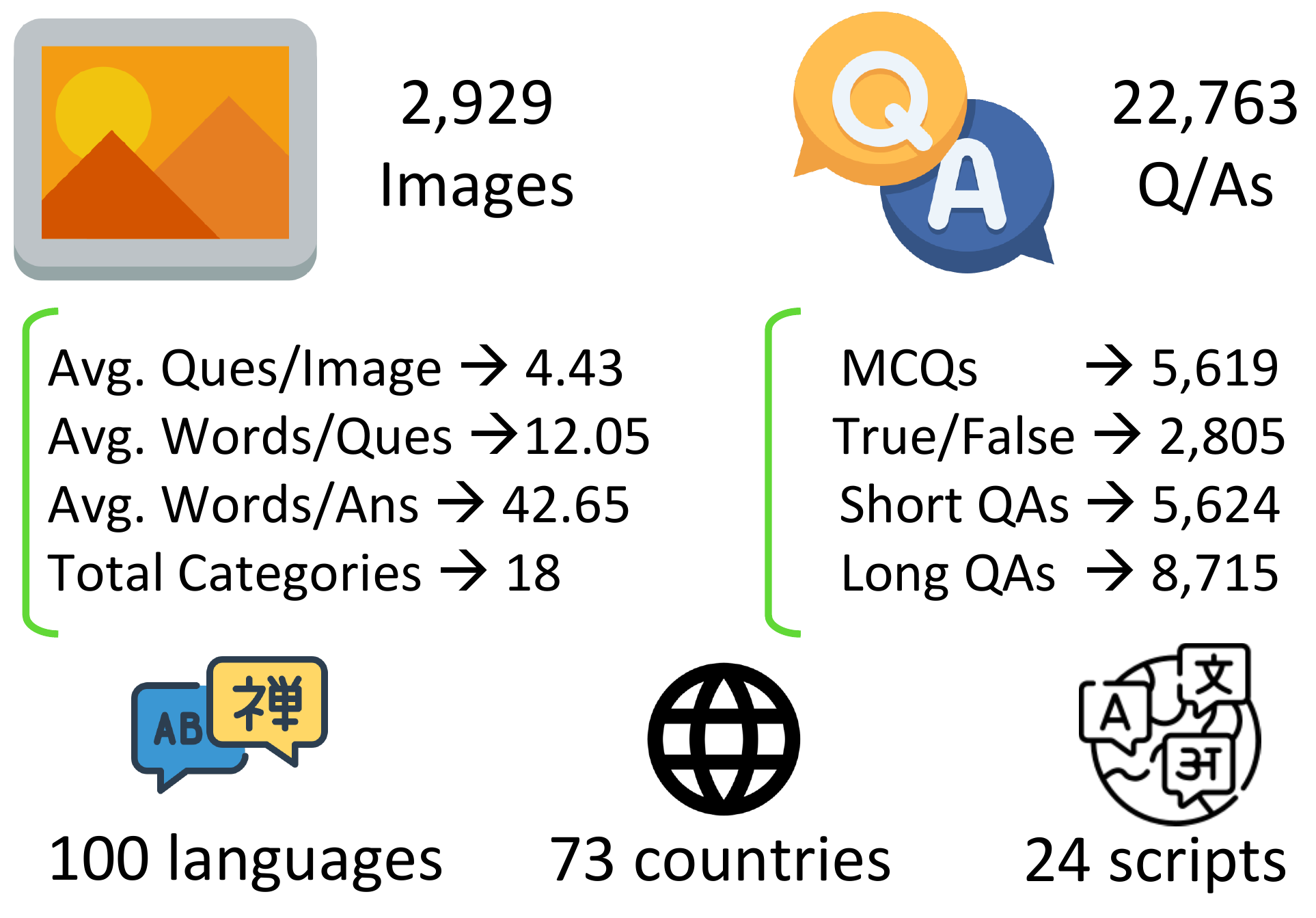}
    \caption{Data statistics of our \ALMbench showing the diversity of the scripts,  global coverage, comprehensive categories, and various question types. Our dataset contains 22.7K high-quality question-answers in total, covering 100 languages and 24 scripts. All the samples are manually verified by native speakers. }
    \label{fig:data_statistics}
    \vspace{-0.5em}
\end{figure}

\ALMbench provides global coverage,  featuring both generic and culture-specific image samples that preserve the nuances of native languages in VQAs (Fig.~\ref{fig:data_statistics}). 
It includes diverse QA formats such as MCQs, True/False, and open-ended VQA (long and short) under 19 domains, grouped into generic and cultural categories Tab. A. 1. (suppl.~material).

The \textit{generic} category includes six domains:  {indoor and outdoor scenes, memes, paintings, food items}, and {sketches} following \cite{liu2024visual}. 
The \emph{cultural} category spans 13 comprehensive domains—Architecture, Customs, Economy, Festivals, Food, Heritage, Lifestyle, Literature, Media, Music, Notable Figures, Religion, and Sports—adding depth to cultural representation across languages.
While some of these domains have been previously explored  \cite{marino2019ok, romero2024cvqa,adilazuarda2024towards, nayak2024benchmarking}, they are investigated in limited languages and do not fully capture the depth of cultural nuances.
Our work introduces six unique domains, including literature, music, festivals, economy, media, and notable key figures to better capture cultural richness across diverse languages. More details are in Sec. A (suppl. material)

\ALMbench comprises of 22,763 diverse questions across all subsets in 100 languages (see Fig. \ref{fig:data_statistics}). 
The dataset spans 73 countries across five continents, capturing the cultural nuances of both underrepresented and predominant languages from different regions. 
We include 24 language scripts (Fig. \ref{fig:scripts_analysis}) and 15 language families, with \emph{Latin} being the most prevalent script, covering 53 languages, followed by \emph{Cyrillic}  with 10 languages; and \emph{Arabic} with 9 languages. In terms of language families, \emph{Indo-European} is the largest group, encompassing 55 languages, with Afro-Asiatic and Austronesian each representing eight languages (Fig. \ref{fig:families_analysis}).

\begin{figure*}[!htp]
    \centering
    \includegraphics[width=\textwidth]{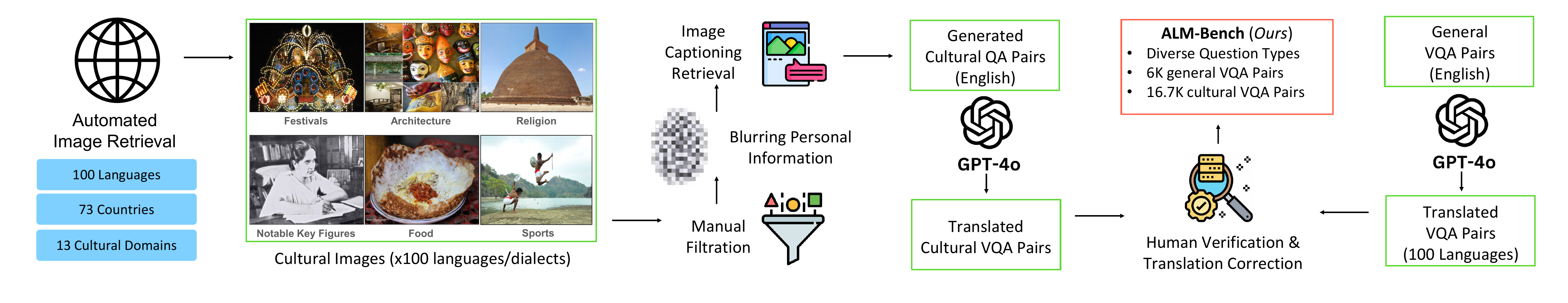}
    \caption{Data collection and verification pipeline. Our benchmark features both cultural specific content sourced from the web (left) and generic image understanding collection sourced from existing LMM benchmark. The cultural part is carefully filtered to remove noisy samples and private information. We use GPT4o for translations which are manually verified and corrected with over 800 hours of human annotators (native speakers). Our \ALMbench has diverse question types and features approximately 23K QA pairs in total in 100 languages.  }
    \label{fig:workflow_diagram}
\end{figure*}

\subsection{Data Collection and Annotation}
\paragraph{Country-Language pairs.}
Our cultural data samples are curated from the country with the most prominent presence of each specific culture.
We carefully selected 100 languages across 73 countries and five continents, striking a balance between low- and high-resource languages and between Western and non-Western cultures to ensure linguistic and cultural diversity.
The selection criteria also consider the number of native speakers and the availability of speakers to validate and annotate the curated samples. 

To accurately capture the cultural nuances of each language, we follow \cite{nayak2024benchmarking} and link each language to a country based on the \textit{World Values Survey} \cite{haerpfer2022world} to cover social, ritualistic, and cultural values. For multilingual countries like India, multiple languages are associated to capture their cultural diversity. For each country-language pair, we collect cultural samples for all 19 generic and cultural domains.

\noindent\textbf{Generic VQA Curation.} 
Our \ALMbench has both generic and cultural questions. 
To evaluate visual understanding in 100 languages on generic scenes, e.g., indoor and outdoor images, we source examples from a well-known English-only LMM benchmark \textit{LLaVA-Bench (In-the-Wild)} \cite{liu2024visual}. These English instructions are extended to the remaining 99 languages using GPT-4o followed by manual correction from native speakers.  In some cases where the GPT-4o translations are poor, the annotators also translate the instructions from scratch. It results in a total of 6,000 open-ended samples across 100 languages for \textit{generic} VQA.

\noindent\textbf{Cultural Image Curation.}
To curate diverse cultural image-QA pairs, we collect open-licensed images and their corresponding metadata from the Internet, focusing on specific country-language pairs across various cultural domains.
For each domain, we go through several filtration steps, e.g., removing low-resolution images and removing redundant samples to ensure diversity.
To maintain high-quality and cultural relevance, we ask expert native speakers of each language to manually verify the quality and cultural diversity of these images. 
Images lacking cultural relevance are removed from the dataset. 
This manual filtration process discarded nearly 10.7\% of the images. 
To ensure privacy, we blur any personally identifiable details (PIDs) and textual information in image that may directly leak the answer. 
Fig.~\ref{fig:workflow_diagram} shows our data collection and verification pipeline.

\noindent\textbf{Cultural QA Generation.} \ALMbench dataset covers different question types. To generate high-quality cultural questions, we utilize GPT-4o to create QA pairs based on the provided images and their metadata. 
Notably, images and QAs are not shared across languages in the cultural set.
For each image, we generate two MCQs, two short questions, a long question, and a true/false question in English. 
We instruct GPT-4o to emphasize the cultural concepts depicted in each image, generate questions that require a visual understanding of an image, and not perpetuate bias and stereotypes. 
To mitigate the randomness in multiple choice and true/false, we ensure that these questions can also be answered when converted to open-ended questions \cite{romero2024cvqa}. We also evaluate with shuffled orders, predicting the entire chosen option rather than just the leading \textit{alphabet}. The detailed instructions are presented in Sec. I (suppl. material.) 

\noindent\textbf{Filtration and Translation Errors.} Given the QA pairs, we translate them to remaining 99 languages using GPT4o, followed by manual correction from native speakers. 
Similar to image filtering, native speakers assess translation quality and make detailed corrections to provide culturally grounded answers focusing on core cultural aspects. 
They also discard redundant and culturally irrelevant QAs. 

\begin{figure}[t]
    \centering
    \includegraphics[width=1\columnwidth]{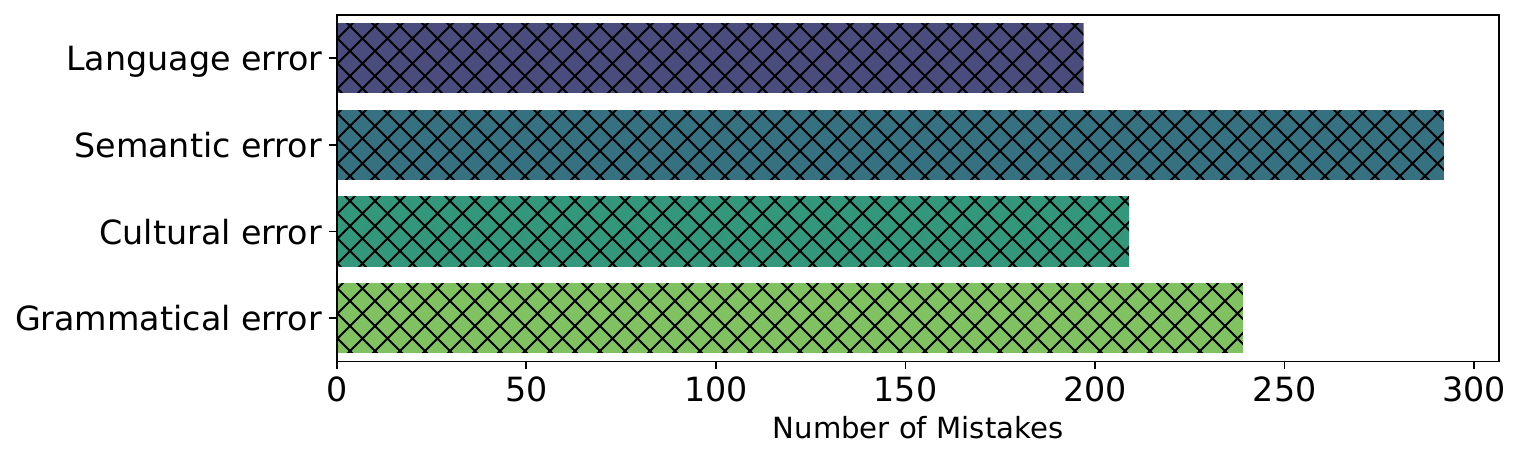}
    \caption{Analysis of the type of mistakes GPT-4o made while language translations via human-feedback. GPT-4o encounters more issues with respect to semantic and grammatical errors.}
    \label{fig:data_annotation}
\end{figure}

We note several challenges in GPT-4o’s translations across 100 languages. Specifically, GPT-4o demonstrates difficulty in generating culturally grounded QA pairs across diverse languages. 
To analyze these issues, mistakes from the GPT-4o model in translations are categorized into four types: \textit{semantic error}, \textit{cultural error}, \textit{language error}, and \textit{grammatical error}. We sample 57 question-answer pairs from 51 randomly selected languages and plotted error distribution in Fig. \ref{fig:data_annotation}. Notably, GPT-4o encounters more issues with semantic and grammatical accuracy when translating into different languages. 
We show examples of such errors in Fig. A. 2. (suppl. material).

\begin{figure*}[!htp]
    \begin{center}
        \includegraphics[width=1\textwidth]{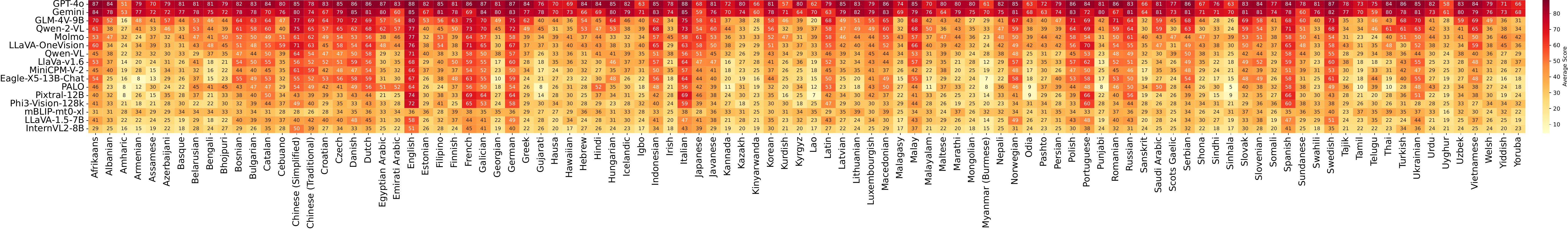}
    \end{center}
    \vspace{-1.5em}
    \caption{\ALMbench Performance comparison of different open and closed-sourced models (y-axis) on the 100 languages (x-axis) of our \ALMbench. The performance is represented as an average accuracy across all questions in a language. The actual performance of a model on a language is shown in each respective box, where the higher accuracy is highlighted with a high color intensity. The closed-source propriety models generally perform better across languages compared to their open-sourced counterparts. The performance on several high-resource languages (e.g., English, French, Chinese, and Spanish) is generally higher throughout different models, whereas all open-source models struggle on low resource languages (e.g.,  Amharic, Kinyarwanda, Burmese)
fs  QAW a     Overall, GPT-4o and GLM-4V-9B performs better in terms of closed-source and open-source models, respectively. Best viewed zoomed in.}
    \label{fig:results_heatmap}
\end{figure*}

\section{Benchmarking LMMs on \ALMbench}

As discussed earlier, \ALMbench comprises four different question types: MCQs, T/F, Short and Long VQA.
Therefore, we employ different prompts for each type of question. 
We use GPT-4o as a judge and prompt it to score responses of various question types with different criterion. We use \emph{accuracy} for MCQs and T/F, \emph{correctness} for short VQAs (SVQAs), and \emph{consistency}, \emph{fluency}, and \emph{relevance} for long VQAs (LVQAs).
Here, correctness refers to how closely the model's output aligns with the ground-truth \cite{liu2024visual}.
For LVQAs, the consistency assesses whether the predicted answer is coherent across the entire generated output \cite{kryściński2019neuraltextsummarizationcritical}.   The fluency metric measures the naturalness and readability of model prediction \cite{sai2022survey},  whereas the relevancy metric determines whether model prediction provides answers directly related to ground truth \cite{chakraborty2024transferqstarprincipled}. To ensure fair evaluation for decision-making questions (T/F and MCQs) using GPT-4o as a judge, we also show consistent performance using LLaMA-3.8.1B-Instruct \cite{dubey2024llama} in Tab. A.2. (suppl. material).

\noindent\textbf{Overall Results:}
Fig. \ref{fig:results_heatmap} presents the per-language performance of 16 recent LMMs on the \ALMbench. 
The results offer a number of insights: \textbf{(a)} the closed-source propriety models (GPT-4o \cite{achiam2023gpt} and Gemini-1.5-Pro \cite{team2023gemini}) generally perform better across the 100 languages, compared to their open-source counterparts. 
The best closed-source model, GPT-4o, achieves an overall accuracy of 78.8\% compared to 51.9\% overall accuracy obtained by the best performing open-source model GLM-4V-9B \cite{glm2024chatglm}.
Both closed-source models struggle with several low-resource languages (e.g., Amharic, Kinyarwanda, Burmese, and Sanskrit). 
For instance, the performance of GPT-4o significantly drops from 88.4\% for the English language to 50.8\% for the Amharic language. 
\textbf{(b)} Several open-source models (GLM-4V-9B \cite{glm2024chatglm}, Qwen2-VL \cite{wang2024qwen2}, Molmo \cite{deitke2024molmopixmoopenweights}, LLaVA-OneVision \cite{li2024llavaonevision}) achieve comparable overall performance. 
Similar to their closed-source counterparts, open-source models also struggle with low-resource languages. 
For instance, the performance of GLM-4V-9B \cite{glm2024chatglm} dramatically drops from 80.3\% for the English language to 15.6\% for the Amharic language. 
Fig. \ref{fig:open_vs_close} presented earlier further shows the overall performance breakdown in terms of high-and low-resource languages across different models. 
It is worth mentioning that our \ALMbench comprises 50 high-resource and 50 low-resource languages following the definition of low-resource languages as in \cite{costa2022no}. 
\textbf{(c)} Fig.~\ref{fig:open_vs_close} reveals a substantial performance gap between high-resource and low-resource languages consistent across different models, with GPT-4o showing a 6\% drop. 
This performance gap extends to more than 8\% in case of some open-source models (e.g., GLM-4V-9B \cite{glm2024chatglm}, Qwen2-VL \cite{wang2024qwen2}). 
A notable exception is the performance of closed-source Gemini-1.5-Pro \cite{team2023gemini}, which does not deteriorate significantly for low-resource languages.
Next, we present further results analysis.

\begin{figure}
    \centering
    \includegraphics[width=0.46\textwidth]{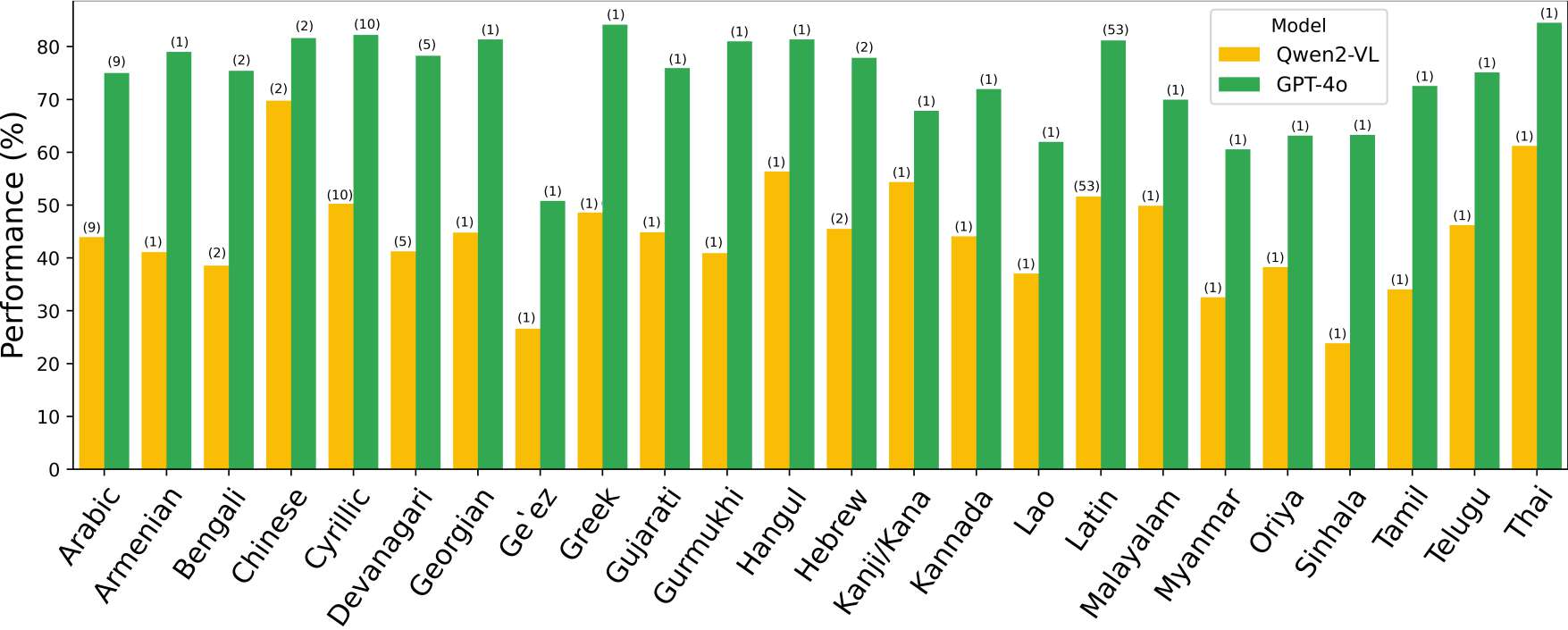}
    \caption{Performance comparison of GPT-4o and Qwen2-VL on different language scripts. The count above each bar indicates the number of languages present in that script.}
    \label{fig:scripts_analysis}
\end{figure}

\begin{figure*}
    \centering
    \includegraphics[width=1\textwidth]{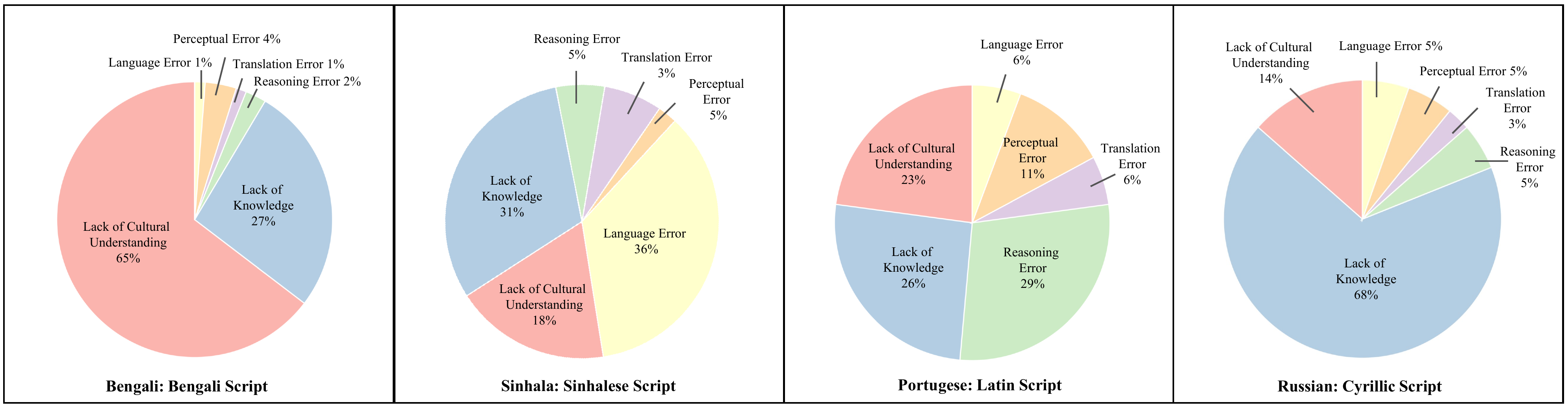}
    \vspace{-1.8em}
    \caption{Error analysis across 4 diverse language scripts, including {Bengali, Sinhalese, Latin and Cyrillic} on GPT-4o results, demonstrates significant challenges for even the top-performing closed-source models, particularly in cultural and reasoning comprehension. The \small\ALMbench highlights these gaps, especially in languages with complex dialectal variations.}
    \label{fig:error_analysis}\vspace{-0.5em}
\end{figure*}

\noindent{\textbf{How important is visual context?}}
To investigate this question, we ran an experiment with only the base LLM of various LMMs used in our evaluation. 
We randomly sample 50 languages and prompt the respective LLM with a textual question without the visual input. Our results demonstrate that the visual context is important to answer several questions in our benchmark and the LLMs struggle without input images. For instance, GPT-4o \cite{achiam2023gpt} performance drops by absolute 27.3\% 
when prompted with textual questions alone. 
Specifically, it shows a significant performance gain in languages such as Sinhala (38.7\%), Sanskrit (50\%), and Dutch (40\%) when images are included.
Similarly, Qwen2-7B gives a 13\% absolute and 24.8\% relative drop in performance without image inputs. 
This substantial performance drop underscores the robustness of our visual benchmark, revealing that without images, even the best-performing proprietary and open-source models LLMs struggle to answer accurately.

\begin{figure}
    \centering
    \includegraphics[width=\columnwidth]{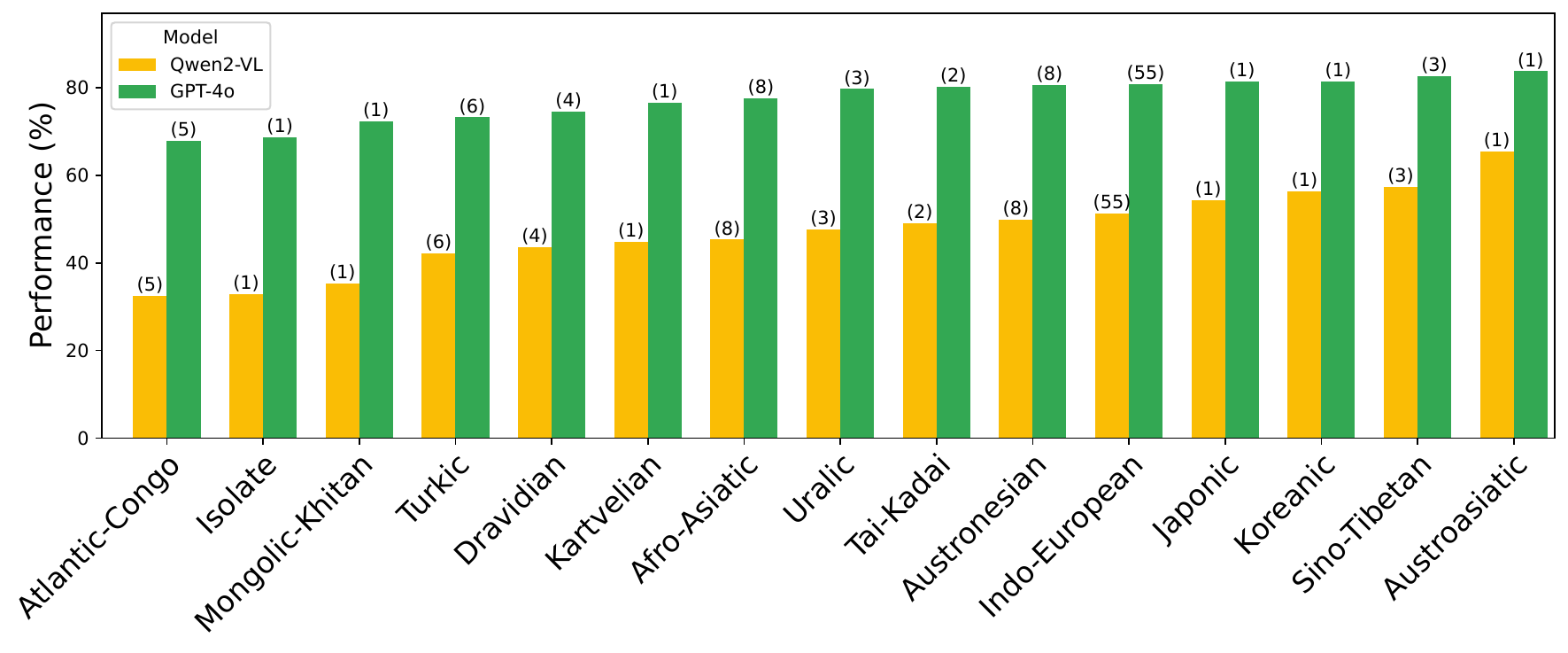}
    \caption{Performance comparison of GPT-4o and Qwen2-VL on different language families. The count above each bar indicates the number of languages present in that script.}
    \label{fig:families_analysis}
\end{figure}

\noindent\textbf{Comparison across
scripts:}
We further group the 100 languages in \ALMbench by script \cite{costa2022no}. We use data from Ethnologue \cite{book} and the Glottolog database \cite{glottolog2022} on language scripts, origins, and families. 
This resulted in 24 distinct script groups.
Fig. \ref{fig:scripts_analysis} shows GPT-4o and Qwen2-VL performance across these scripts, with both models struggling particularly on low-resource scripts such as,  \textit{Ge'ez} (Amharic), \textit{Sinhalese} (Sinhala), \textit{Oriya} (Odia), and \textit{Myanmar} (Myanmar-Burmese). 
While the results are not surprising as they highlight that existing models are being predominantly trained on high-resource languages, the extent of decline for low-resource scripts is noteworthy. 

\begin{figure*}
    \centering
    \includegraphics[width=1\textwidth]{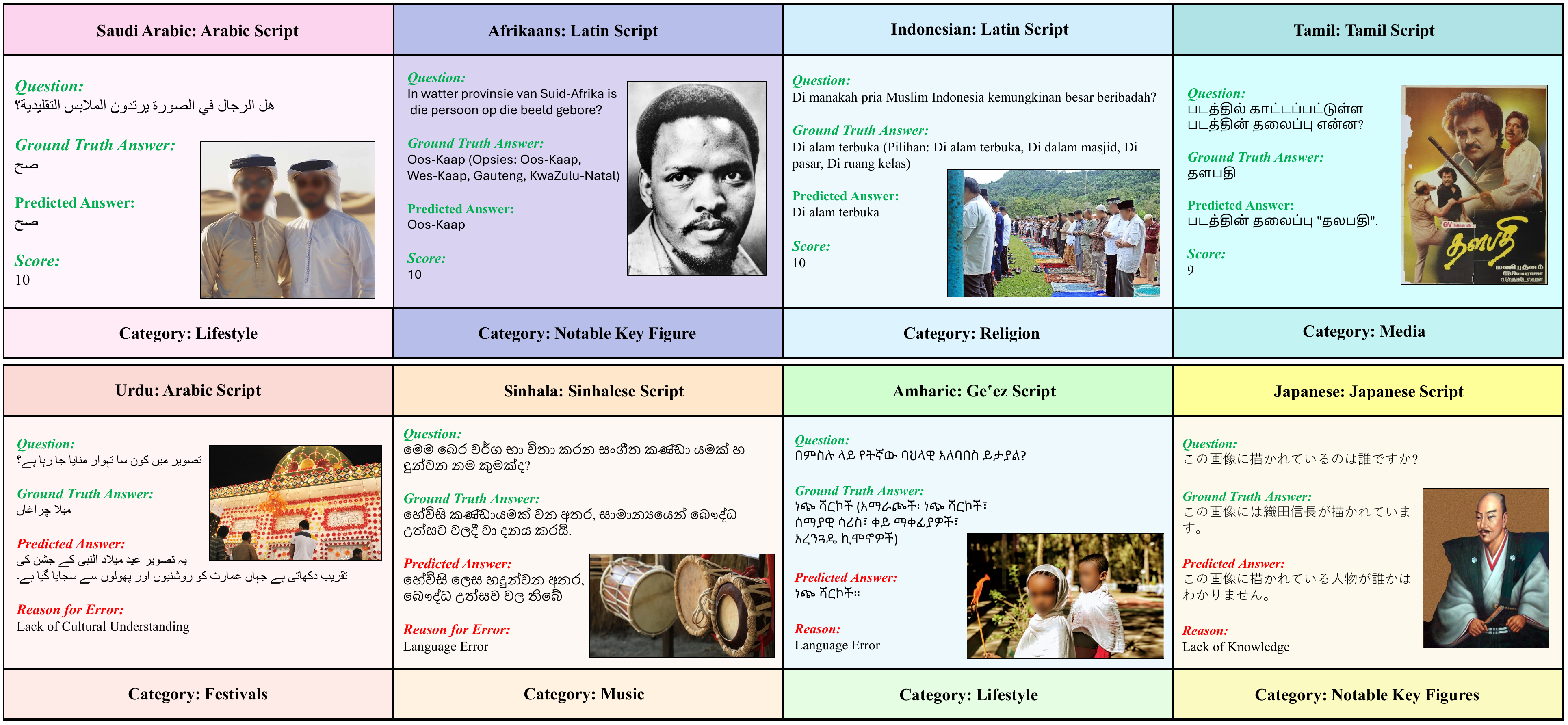}
    
    \caption{We present the qualitative examples of the success cases in the first row and failure cases of GPT-4o in the second row on different languages \& domains in \ALMbench. For the failure cases, we specify different error types. 
    For instance, the Urdu language question asks about the festival depicted in the image. The image specifically refers to \emph{Mela Chiraghan} (Festival of Lights), a celebration held in honor of the Sufi saint Shah Jamal’s shrine. Since the decoration in the image closely resembles that of Eid Milad un Nabi — another religious festival—the model erroneously associates it with this wrong event. This constitutes a lack of cultural understanding since the model fails to distinguish between the theme behind the decorations. Eid Milad un Nabi typically features more modest, reverential lighting with green lights, whereas the lighting in Mela Chiraghan is brighter and more colorful. Additionally, people typically dress for the Eid Milad un Nabi event in a traditional outfit which is absent in the image. These examples highlight the model’s gap in cultural knowledge and its limitations in terms of accurately interpreting the cultural context of the given sample. Additional results are in the suppl. material.}
    \label{fig:qual_error_analysis}
\end{figure*}

We conduct an error analysis on the cultural examples from \ALMbench by selecting one language per script and having native speakers review GPT-4o (best-performing model) responses. 
Errors were categorized into six types: lack of knowledge, reasoning error, perceptual error \cite{yue2024mmmu}, language error, translation error, and lack of cultural understanding. 
 Fig.~\ref{fig:error_analysis} summarizes these error types across four scripts, showing that the primary issues involve knowledge gaps, reasoning errors, and cultural misunderstandings.
Fig.~\ref{fig:qual_error_analysis} further illustrates specific errors across scripts. 
One such example in Sinhala asks about the type of music group that uses two traditional drums, the \emph{Thammattama} and \emph{Dhawula}, which are commonly played by \emph{Hewisi} music groups associated with traditional Buddhist festivals.
Although GPT-4o correctly answered \emph{Hewisi} and recognized its role in Buddhist festivals, the response is incorrectly written in the Sinhala language. This highlights GPT-4o's language error with the Sinhalaese script, demonstrating a gap in script proficiency despite grasping cultural context.

\noindent\textbf{Comparison across language families:}
We also analyze the performance of LMMs by grouping the different languages with respect to 15 families taken from Britannica\footnote{https://www.britannica.com/browse/Languages} as shown in Fig. \ref{fig:families_analysis}. 
Results show that performance on several African (\textit{Atlantic-Congo}) languages such as Igbo, Kinyarwanda, Shona, Swahili, and Yoruba is inferior compared to several Asian (e.g., Chinese, Korean, Vietnamese) and Western languages (e.g., English, French, German). 

\begin{figure}
    \centering
    \includegraphics[width=\columnwidth]{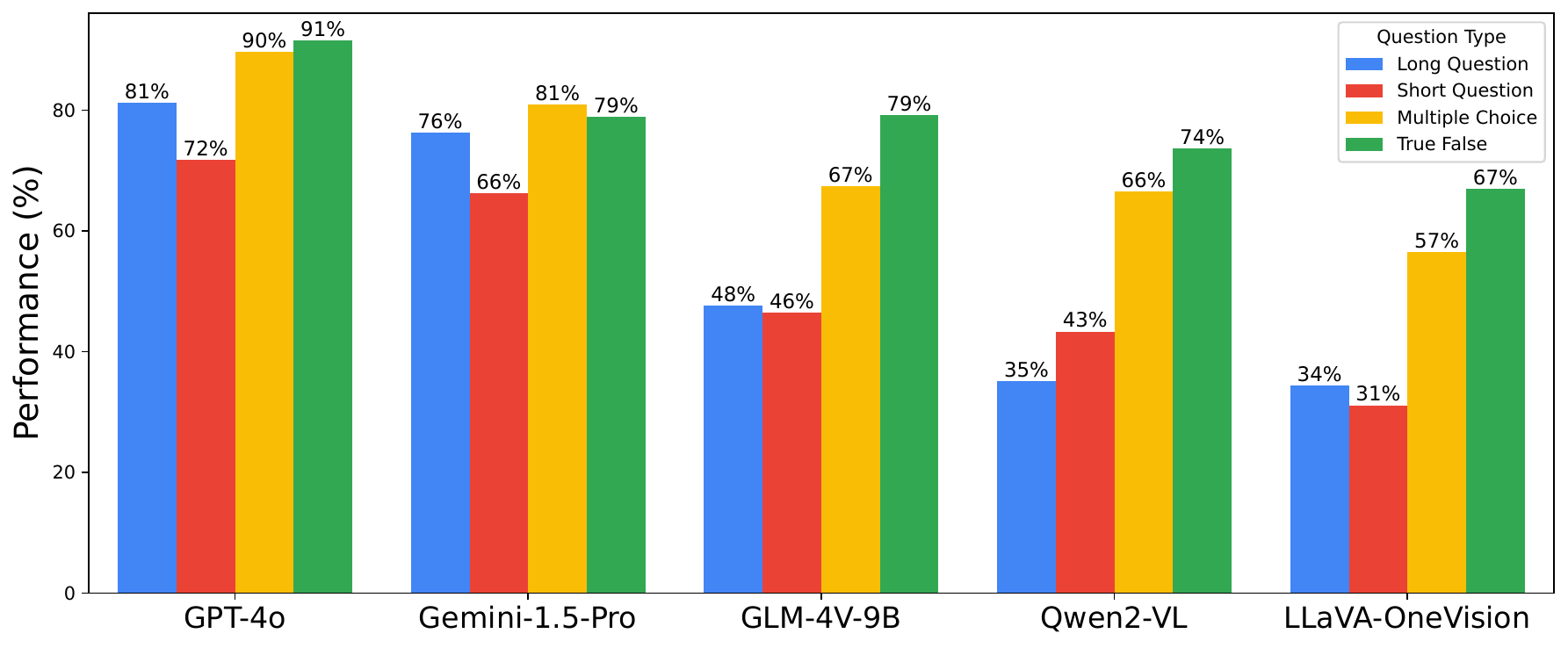}\vspace{-0.3em}
    \caption{Performance of top-performing VLMs on \ALMbench according to different question types. Compared to open-ended questions, VLMs perform better on MCQs and T/F questions.}
    \label{fig:question_types}
\end{figure}

\noindent\textbf{Effect of question types:} We perform this analysis on two closed-source (GPT-4o \cite{achiam2023gpt} and Gemini-1.5-Pro \cite{team2023gemini}) and two open-source (GLM-4V-9B \cite{glm2024chatglm}, Qwen2-VL \cite{wang2024qwen2}) models, as shown in Fig.~\ref{fig:question_types}. Overall, we observe that all models fare better on decision-making questions (MCQs and T/F questions). Closed-source models perform better on long VQAs (LVQAs) than short VQAs (SVQA). We observe this trend to be the opposite in open-source models as they struggle to generate accurate and fluent long responses in a multilingual setting with 100 languages.

\noindent\textbf{Are LMMs Culturally Aware?} We study the cultural understanding of LMMs on our 13 domains and report results in Fig. \ref{fig:category_wise}. 
Overall, GPT-4o achieves the best results with an overall score of 80.3\%. 
However, we observe a significant variation in performance across different cultural domains. 
For instance, GPT-4o achieves 83.7\% on \textit{Education and heritage} but drops to 72.7\% on \textit{Notable key figures} category. 
We conjecture this variation in performance likely arises since the higher-performing categories are covered in LMM training datasets, leading to better representation learning. 
In contrast, categories like \textit{Notable Key Figures} and \textit{Customs} are often culturally specific and nuanced across languages and regions.

\noindent\textbf{Impact of location-aware information in prompts:} 
We show the performance of LMMs when evaluated with additional country-specific information in Tab. \ref{tab:country_info_performance}. We randomly select 50 languages from the \ALMbench and include country information in the prompts to assess any potential performance gains. Here, we observe that closed-source models, including GPT-4o and Gemini-1.5-Pro, better utilize the added geographic context and indicate higher cultural specificity across languages than the open-source models.

\begin{table}
    \centering
    \resizebox{\columnwidth}{!}{%
    \begin{tabular}{lcc}
        \toprule
        \textbf{Models} & \textbf{With Country Info.} & \textbf{Without Country Info.} \\
        
        \midrule

        \rowcolor{LGray} GPT-4o & \textcolor{darkgreen}{83.57\%} & 80.96\% \\
        Gemini-1.5-Pro & \textcolor{darkgreen}{81.52\%} & 76.19\% \\

        \arrayrulecolor{black}
        \cdashline{1-3}[2pt/2.5pt]
        
        \rowcolor{LGray} GLM-4V-9B & \textcolor{darkgreen}{56.78\%} & 56.41\% \\ 
        Qwen2-VL & \textcolor{darkgreen}{53.97\%} & 52.57\% \\
        
        \bottomrule
    \end{tabular}
    }
    \caption{Performance of LMMs with and without additional country location information. Proprietary models show a notable performance boost of 2.6\% to 5\% when location-aware prompts are used, while open-source models exhibit a marginal improvement.}
    \label{tab:country_info_performance}
\end{table}
\section{Ethical Consideration}

\textbf{Data.} Our cultural images in \ALMbench are sourced from the internet. Therefore, there may be potential biases in underrepresented domains. However, we believe \ALMbench represents a step forward in standardized multilingual-LMM evaluation, paving the way for greater inclusion of diverse cultural domains, languages, scripts, and origins. \\
\textbf{Translations.} We use GPT-4o \cite{achiam2023gpt} for translations in 100 languages. The initial translations generated by GPT-4o undergo a thorough review by native speakers to ensure linguistic accuracy and cultural relevance to maintain high-quality standards across all language pairs. This combined approach enables consistent and accurate translations. \\
\textbf{Human Annotations.} To ensure high-quality image filtering and language annotation, we collaborated with linguistic groups and finalized over 60 volunteers for benchmark curation. 
For most languages, annotators were required to be either native or bilingual speakers. However, for certain low-resource languages like Igbo, Hungarian, and Afrikaans, proficiency was also acceptable. Annotators were required to be familiar with the cultural context of the specific country-language pair they were working with.  
Demographic details about the expert annotators are provided in Sec. B (suppl. material). Among them, 80.3\% are native speakers, and 87.9\% have lived for over 15 years in countries where the target language is spoken, thereby ensuring a deeper cultural insight. 
Furthermore, 46.7\% of annotators both male and female, fall within the 18–25 age bracket.
\section{Conclusion}
In this paper, we introduce \ALMbench, a novel multilingual multimodal cultural benchmark for evaluation with over 22.7k humanly verified samples across 19 domains. Our benchmark encompasses cultural nuances from 73 countries in 24 language scripts and 15 language families. We conduct empirical analysis on 16 vision-language models with various question types (MCQs, T/F, SVQA, and LVQAs) and highlight notable disparities in their performance.

The performance difference between the best-performing open-source model and the proprietary model, GPT-4o, is 27\%. Our results also highlight that the models perform superior on predominant language scripts such as \textit{Latin, Cyrillic}, and \textit{Devanagari} and under-performs on underrepresented scripts such as \textit{Ge'ez, Lao, Sinhalese}, and \textit{Oriya}. Moreover, cultural understanding of prominent language families such as \emph{Indo-European}, 
\emph{Austronesian} and \emph{Afro-Asiatic} are well represented by GPT-4o as compared to \emph{Atlantic-Congo} and \emph{Turkic} families. Our work highlights the limitations of state-of-the-art LMMs in multilingual and multicultural settings, showing key areas for improvement.

\section{Acknowledgment}
We would like to thank Tafar Mab for providing high-quality annotations for Afrikaans and Kinyarwanda languages, Georgios Ioannides for verifying Greek language, Marek Suppa for annotating and verifying Czech, Slovak and Polish language, and Feno Heriniaina Rabevohitra for annotating Malagasy language. We also thank Yuhao Li for reviewing and assisting with the annotations for Chinese Simplified language. Finally, we extend our gratitude to the anonymous reviewers for their invaluable suggestions and feedback, which helped improve this paper. The computations were enabled by resources provided by NAISS at Alvis partially funded by Swedish Research Council through grant agreement no. 2022-06725, LUMI hosted by CSC (Finland) and LUMI consortium, and by Berzelius resource provided by the Knut and Alice Wallenberg Foundation at the NSC. This work was partially supported by by the Swedish Research Council (2022-04266) and from KAW (DarkTree project; 2024.0076).

{
    \small
    \bibliographystyle{ieeenat_fullname}
    \bibliography{main}
}



\clearpage



\renewcommand{\thefigure}{A.\arabic{figure}} %
\setcounter{figure}{0} 
\renewcommand{\thetable}{A.\arabic{table}}
\setcounter{table}{0} 
\renewcommand{\thesection}{\Alph{section}} %
\setcounter{section}{0}


\appendix

\section*{Appendix}

\section{ALM-Bench Categories}
\label{sec:ALM-Bench_Categories_Appendix}
Our \ALMbench benchmark dataset consists of 19 categories. Among these categories, we include 
\textit{Food, Lifestyle, Religion, Architecture, Sports, Customs} and \textit{Heritage} from \cite{romero2024cvqa, marino2019ok} with some minor modifications to it. We further include six additional categories including \textit{Literature}, featuring prominent authors, poets, and their notable works; \textit{Music}, showcasing traditional music and dance through visual samples; \textit{Festivals}, covering major cultural celebrations; \textit{Economy}, representing local industries and businesses; \textit{Media}, highlighting cultural icons, entertainment figures, and popular TV shows; and \textit{Notable Key Figures}, featuring influential historical leaders who serve as representatives for country-language association. Table \textcolor{red}{A.} \ref{tab:categories} entails all our \ALMbench categories.

\begin{figure}
    \centering
    \includegraphics[width=0.4\textwidth]{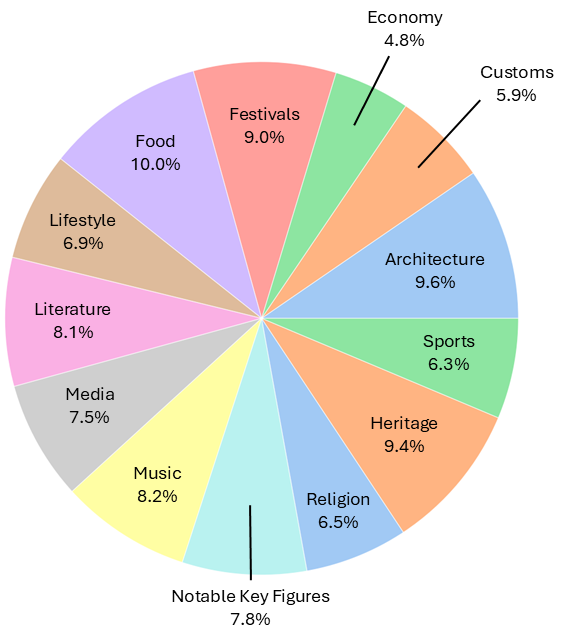}
    \caption{A breakdown of the cultural categories from our \ALMbench is depicted. We ensure consistent samples across all subsets, except for the \textit{Economy} category, where culturally unique images were challenging to find. These visual and question-answer samples are verified and filtered by the native speakers, removing any culturally irrelevant and redundant information.}
    \label{fig:categories_distribution}
\end{figure}

\begin{figure*}[t!]
    \centering
    \includegraphics[width=1\textwidth]{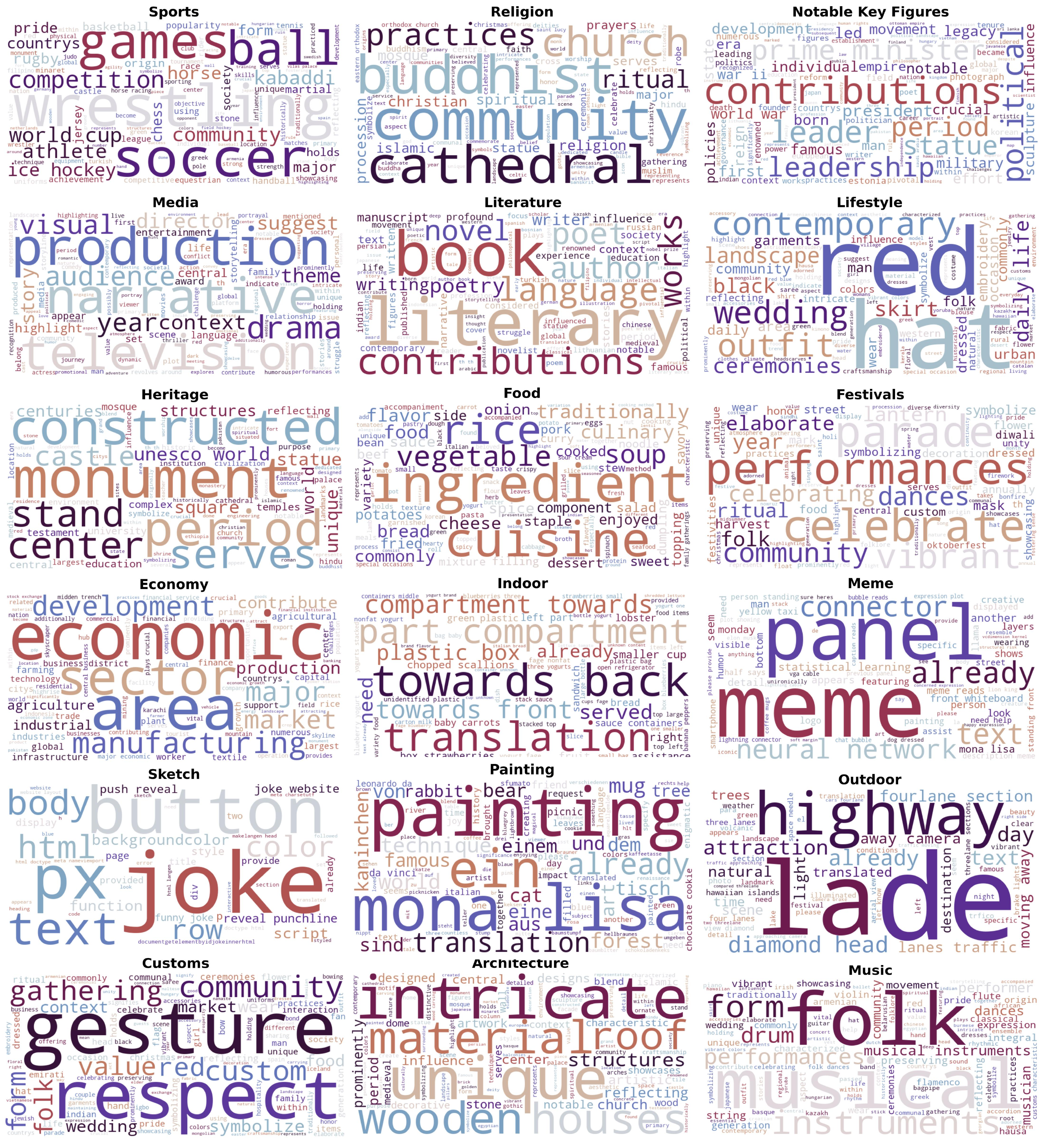}
    \caption{Word clouds depicting prominent concepts from 19 categories in our \ALMbench. For intriguing results, we plot and demonstrate the results on English samples of the plot representing both the cultural and generic elements for the entire 100 languages.}
    \label{fig:word_plot}
\end{figure*}

Following \cite{adilazuarda2024towards}, we adopt an approach to group cultural attributes based on the country for each language. Additionally, we group culturally grounded elements representing shared knowledge, values, and objectives among the people in that culture that are collectively understood. We curated a culturally diverse dataset by collecting data from many cultural aspects (see Fig. \textcolor{red}{A.} \ref{fig:categories_distribution}) and manually curated the caption for each image from web-sources. We also show the most frequent words from these categories in Fig. \textcolor{red}{A.} \ref{fig:word_plot}.

\begin{table}[H]
\centering
\resizebox{0.4\textwidth}{!}{
\begin{tabular}{@{}ll@{}}
\toprule
\textbf{\texttt{ALM-bench} Categories} \\ \midrule
\begin{tabular}[t]{@{}l@{}}
1. Indoor \\
2. Outdoor \\
3. Food Items \\
4. Memes \\
5. Painting \\
6. Sketch \\
7. Food \\
8. Lifestyle \\
9. Religion \\
10. Literature \\
\end{tabular} &
\begin{tabular}[t]{@{}l@{}}
11. Music \\
12. Customs \\
13. Festivals \\
14. Heritage \\
15. Economy \\
16. Media \\
17. Architecture \\
18. Sports \\
19. Notable Key Figures \\
\end{tabular} \\ 
\bottomrule
\end{tabular}
}
\vspace{-0.1in}
\caption{The 19 categories present in our \texttt{ALM-bench} dataset. Note that \textit{Food Items} and \textit{Food} appear in both generic and cultural categories, but have a different context as described in Sec. \ref{box:cultural-domains}.}
\vspace{-0.1in}
\label{tab:categories}
\end{table}

\section{Annotators Demographics}
\label{sec:annotators_demographics}
Our \ALMbench is the most comprehensive benchmark to date, capturing cultural nuances across 100 languages. Since building such large-scale datasets is challenging, especially due to the limited availability of high-quality native language samples, we enlisted over 60 volunteers to provide expert feedback on our curated examples. In addition to our in-house team, we collaborated with multilingual research communities to bring in external volunteers. These volunteers, representing 50 different countries, bring a deep understanding of their languages’ cultural elements and insights unique to their country-language pairs. 

\begin{figure}[H]
    \centering
    \includegraphics[scale=0.2, width=\columnwidth]{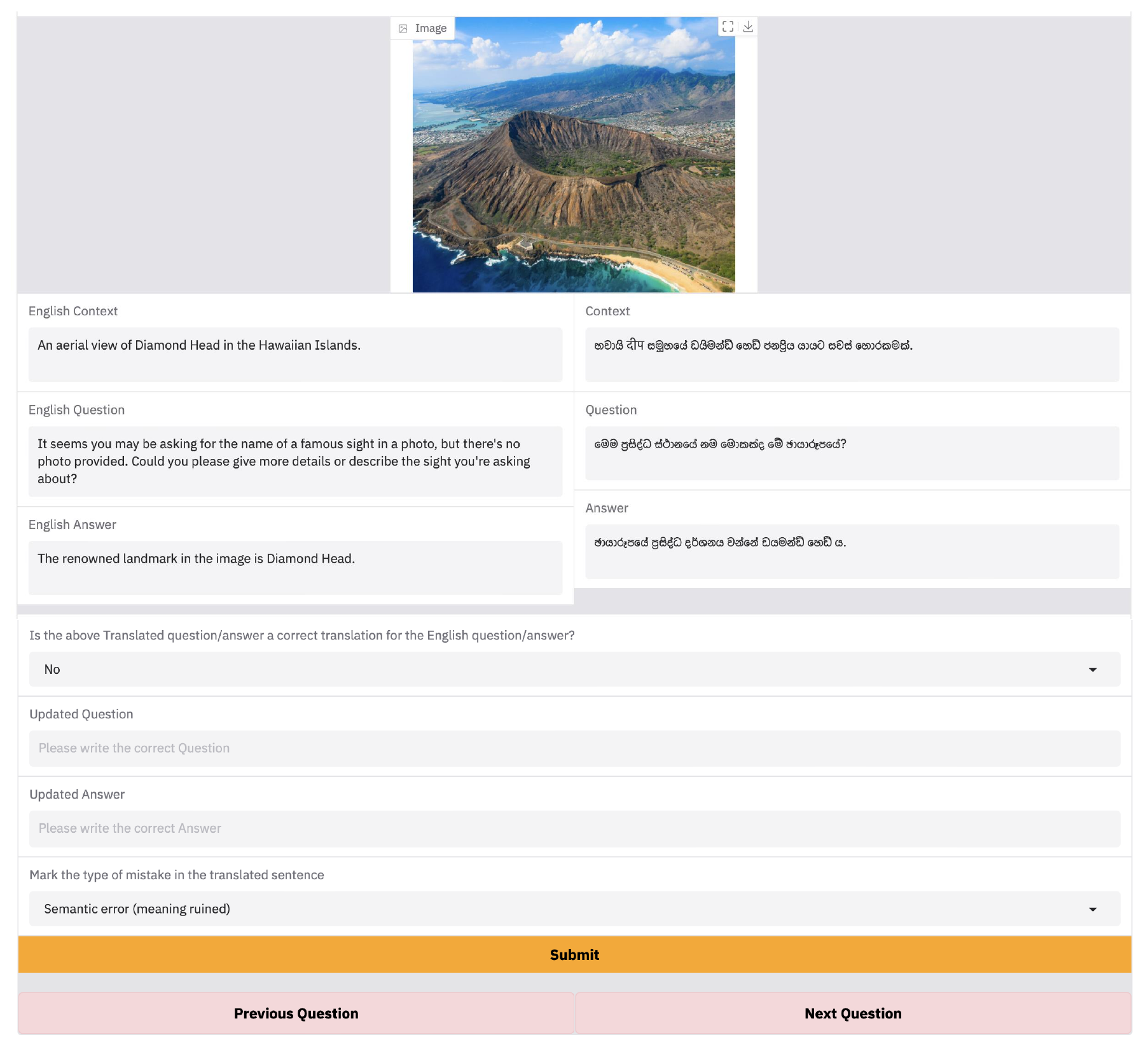}
    \caption{User interface for translation verification hosted on Gradio, allowing contributors to classify incorrect entries and provide accurate translations.}
    \label{fig:UI_examples}
\end{figure}

Each annotator was given detailed instructions along with examples to guide them in curating high-quality datasets. We also hosted a UI interface as well as gave them the option to directly use the dataset. An example of our annotation platform is shown in Fig. \textcolor{red}{A.} \ref{fig:UI_examples}. Annotators who made substantial contributions are recognized as co-authors of this paper to support high-quality language verification.

\begin{figure}[ht]
    \centering
    \includegraphics[width=0.4\textwidth]{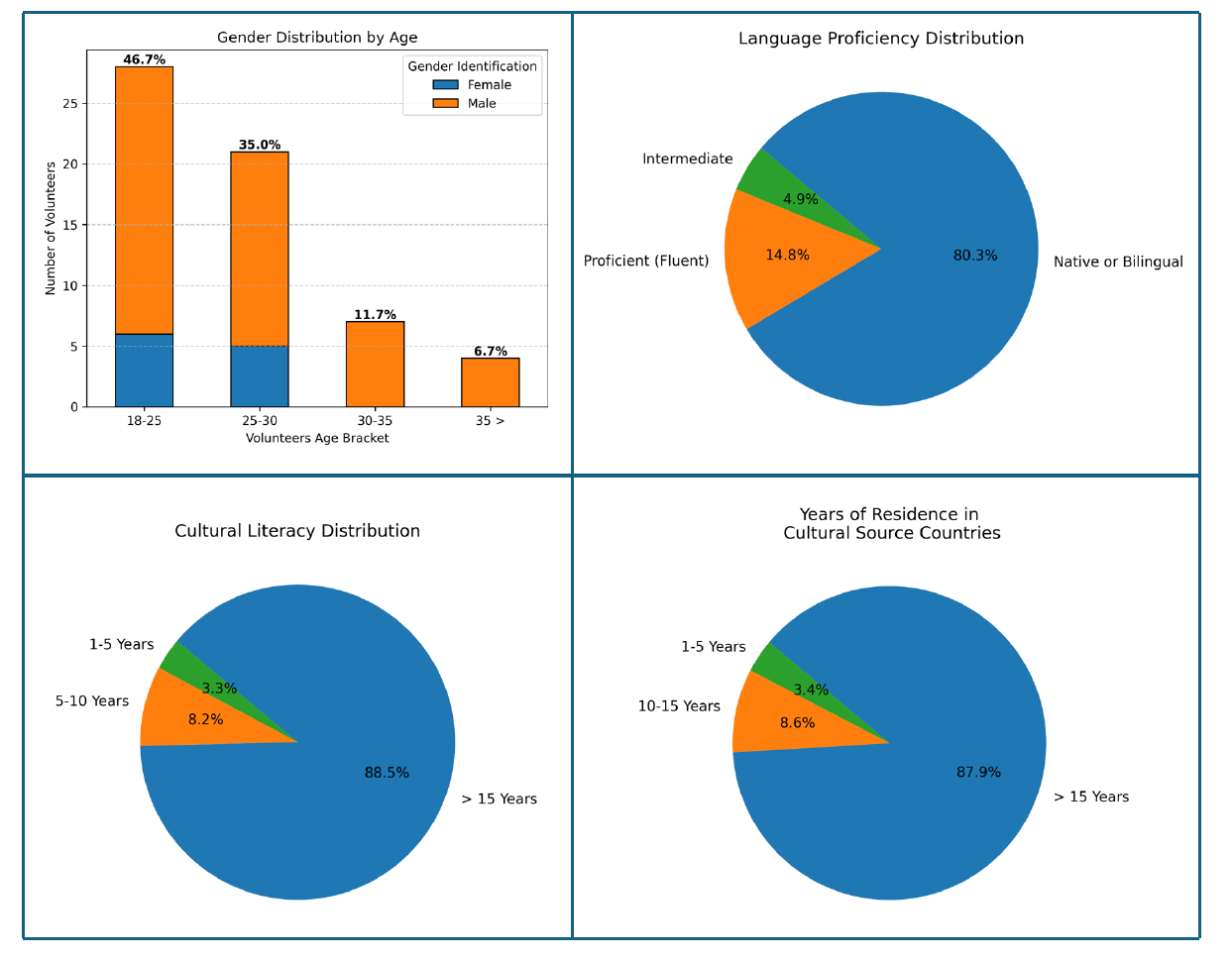}
    \caption{\textbf{top-left.} depicts the \emph{Age} and \emph{Gender}  distribution of our volunteers. We have one-fourth of the Female speakers and it is also shown that almost 48.3\% of our volunteers fall in the age-bracket of 18-25. \textbf{top-right.} depicts the volunteer's native language proficiency in years. \textbf{bottom-left.} depicts the volunteer's Cultural Literacy proficiency in years. \textbf{bottom-right.} depicts the volunteer's native language proficiency in years.}

    \label{ref:annotator_demographics}
\end{figure}

We summarize the annotator demographics in Fig. \textcolor{red}{A.} \ref{ref:annotator_demographics}. To summarize, we show the Age and Gender distribution of our volunteers (\textit{top-left}) indicating the presence of one-fourth of Female annotators and over 46\% fall in the 18-25 age limit. We show their levels of language proficiency (\textit{top-right}). Notably, over 80\% of the annotators were either native speakers or bilingual, with an additional 14.8\% at a proficient level in their chosen language. This high level of expertise ensures that our cultural samples have been meticulously reviewed, filtered, and refined by knowledgeable volunteers. Additionally, we pre-select annotators who are accustomed to the cultural values of their languages. Moreover, (\textit{bottom-left}) depicts that over 88.5\% are culturally familiar with their languages, and \textit{bottom-right} highlights that 87.9\% of the annotators have lived in the country where we have crafted the cultural samples, making them ideal participants for participating in this research effort.

\begin{figure}[t]
    \centering
    \includegraphics[scale=0.2, width=\columnwidth]{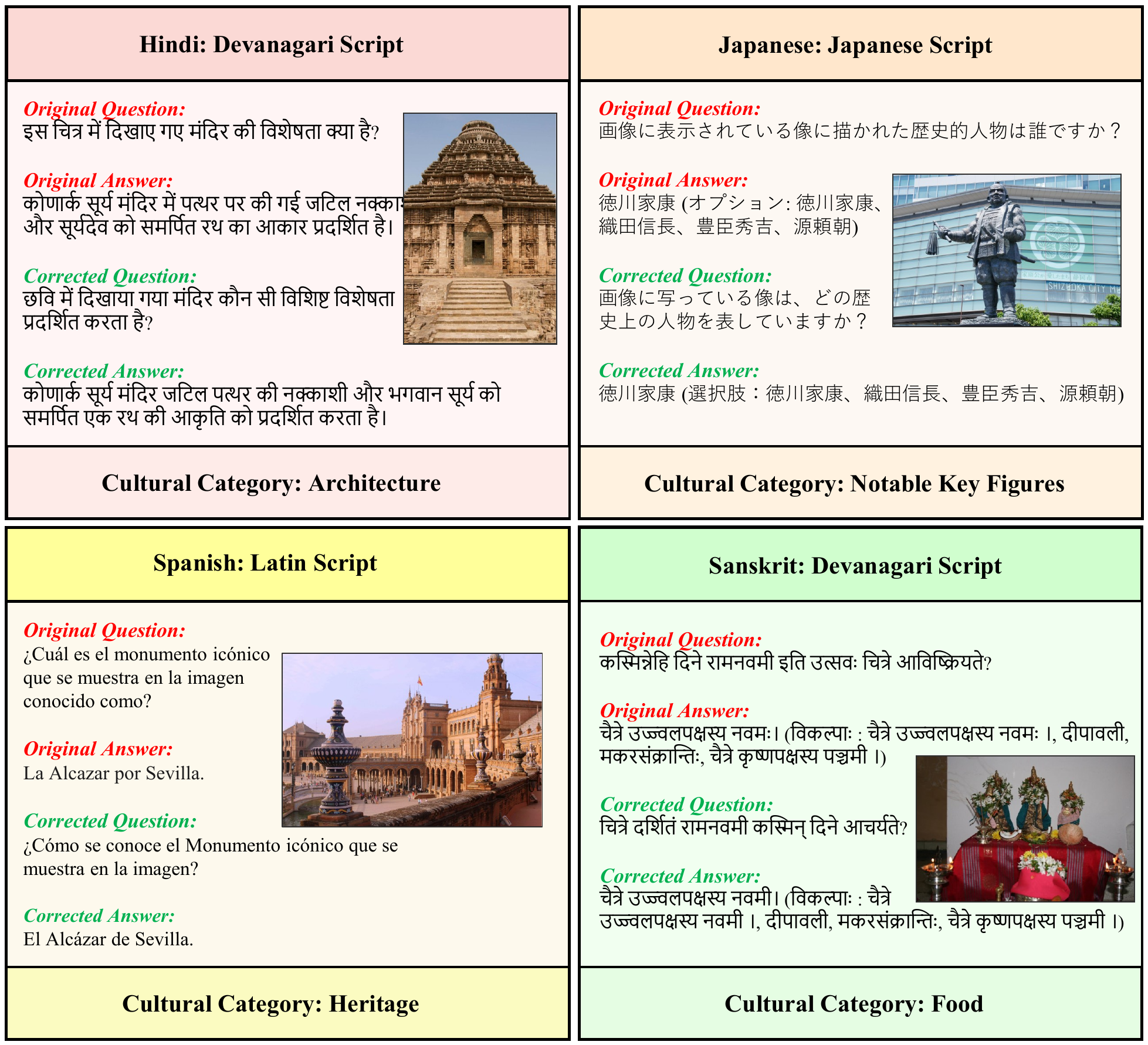}
    \caption{Qualitative examples of various mistakes in GPT-4o translations including \textit{semantic, cultural, language and grammatical errors}. We employ expert human-feedback to rewrite the correct translations for all samples in our \ALMbench dataset.}
    \label{fig:translation_correction}
\end{figure}

\section{Guidelines for Data Verification}

For each country-language pair (e.g., Sri Lanka-Sinhala), QA pairs for general domains are created by translating the original English QA pairs from \textit{LLaVA-Bench (In-the-Wild)} dataset \cite{liu2024visual}. For cultural domains, we curate the dataset by collecting images from online sources using targeted keywords (e.g., country name, language name, cultural category) for each language. We also fetch meta-data along with the image, including image caption, image size, and image license. We use such information for post-processing to keep only high-quality images. Then, we perform manual filtering and blurring of personally identifiable data (PID). We discuss this step in detail in Sec. \ref{sec:blurr_PIDs}.  The cleaned images are used to generate and translate QA pairs using GPT-4o. To ensure quality, contributors with expertise in the respective native languages manually verify the cultural relevance of images, as well as the accuracy, relevance, and consistent translation of the generated QAs. This section outlines the instructions provided to contributors for verifying the relevance of questions generated for cultural domains and for reviewing the QA translations across both general and cultural domains. Additionally, we use GPT-4o as a judge to score the response and assess the accuracy of answers for cultural category images. This process is detailed further in Sec. \ref{sec:GPT_Scoring} and \ref{sec:Error_Analysis}.

Each annotator is provided access to a dedicated verification platform that guides them through the list of QA pairs for the selected language (examples are presented in Sec \ref{sec:verification_platform}). Each QA pair includes the following components: Image, Category, English Question, English Answer, Translated Question, and Translated Answer. Contributors are required to follow the two steps outlined below.

\textbf{\textcolor{blue}{Step 1}} (Question Relevance): Verify that the question is phrased such that it cannot be answered without the use of a Vision-Language Model (VLM) analyzing the image. For example, a question like “Where is Aura Church located?” can be answered without any visual input. A more appropriate question would be, “Where is the church shown in the image located?"

\textbf{\textcolor{blue}{Step 2}} (Translation Verification): If the translated question or answer is incorrect, provide the correct version in the designated space and classify the error into one of the following four categories.

\begin{tcolorbox}[colback=blue!5!white, colframe=blue!75!black, title=Translation Error Types]

\begin{itemize}
    \item \textbf{Semantic error}: Translation hasn't captured the semantic meaning 
    \item \textbf{Cultural error}: Correct but unusual words uncommon in local language context are used
    \item \textbf{Language error}: Translation is provided using characters from a different alphabet or most commonly not translated and still given in English
    \item \textbf{Grammatical error}: Grammar-related errors present
\end{itemize}

\end{tcolorbox}

\section{Verification Platform and UI Examples}
\label{sec:verification_platform}

\section{Image Web Searching}

We automated the process of web scraping for collecting cultural images. A custom URL, embedded with our encoded search query and licensing information, is passed to the Google Search Engine, and images are downloaded from the retrieved links. For each specific cultural domain within a given country-language pair, the search query is constructed as ``\texttt{\{language\} \{cultural domain description\} in \{country\}}''. For privacy and copyright issues, our benchmark only includes images that are either in the public domain or licensed under the Creative Commons license.

\section{Blurring the PIDs}
\label{sec:blurr_PIDs}
Recent research highlights the importance of blurring personally identifiable details (PIDs) in image datasets to mitigate privacy risks and reduce bias \cite{zhang2024discover}. Thus, we filter the collected cultural images for cultural relevance, and faces are blurred using a face detection model, except for the ‘Media’ and ‘Notable Key Figures’ categories where we have images of public figures or celebrities. The blurred images are then manually reviewed to identify any discrepancies, such as blurring personally identifiable data (PIDs) and removing watermark images, and any remaining issues are addressed when using the \href{https://picdefacer.com/en/}{PicdeFacer} tool.

\begin{tcolorbox}[colback=blue!5!white, colframe=blue!75!black, title=Cultural Domain Descriptions, label=box:cultural-domains]

\begin{itemize}
    \item \textbf{Food}: Specific Dishes, Dining Dishes
    \item \textbf{Lifestyle}: Daily Attire, Daily Life, Modern Lifestyle
    \item \textbf{Literature}: Famous literature, authors and poets
    \item \textbf{Music}: Traditional music and dance
    \item \textbf{Religion}: Major religions and religious festivals
    \item \textbf{Customs}: Social etiquette and traditional greetings
    \item \textbf{Festivals}: Important cultural festivals and celebrations
    \item \textbf{Heritage}: Popular historical heritage sites and iconic landmarks
    \item \textbf{Economy}: Major economic industries
    \item \textbf{Media}: Iconic Entertainment Figures and Popular TV Shows, Movies
    \item \textbf{Architecture}: Traditional Art and Architecture
    \item \textbf{Sports}: Famous traditional sports
    \item \textbf{Notable key figures}: Famous historical leaders
\end{itemize}

\end{tcolorbox}


\section{Instruction Prompt for QA Generation}

\begin{figure*}[t!]
    \centering
    \includegraphics[width=1\textwidth]{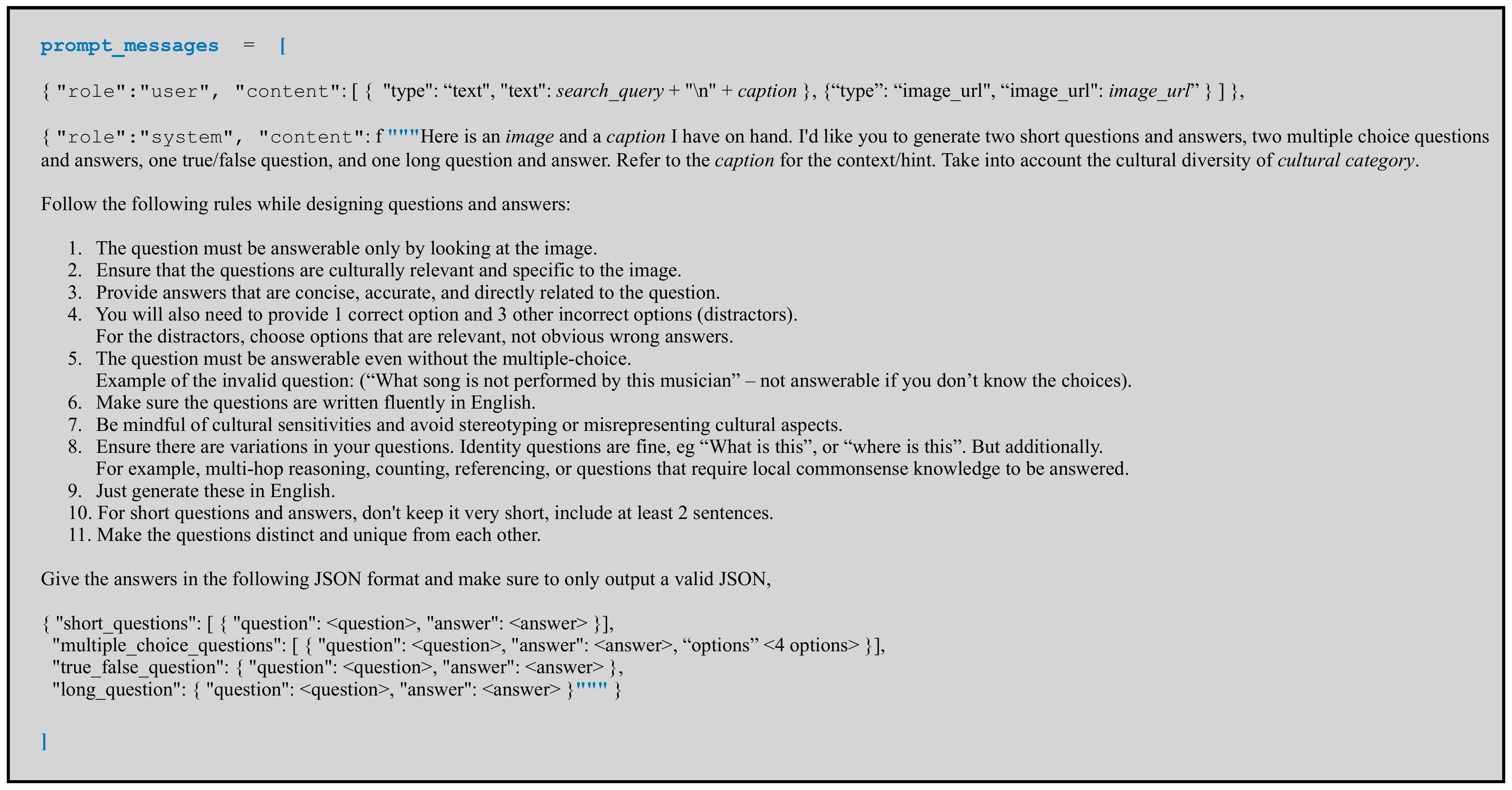}
    \caption{Prompts used for generation of cultural QA pairs. \textit{search\_query} refers to the query used to search the image online. It includes language, country and cultural category details. \textit{caption} is a manually added textual description specific to the image. \textit{cultural category} indicates the domain to which the image belongs, selected from the 13 cultural domains in our \ALMbench dataset.}
    \label{fig:QAgen_prompts}
\end{figure*}

Fig. \textcolor{red}{A.} \ref{fig:QAgen_prompts} illustrates the prompt messages used to guide GPT-4o in generating culturally relevant QA pairs. Each prompt includes an image, its manually added caption, and the search query used during image scraping. The model is instructed to generate two short QA pairs, two multiple-choice QA pairs, one true/false question, and one long QA pair. Detailed instructions ensure that the generated QAs are culturally grounded and specific to the image, avoiding answers based solely on common knowledge.

\section{Instruction Prompt for QA Translations}
\begin{figure}[H]
    \centering
    \includegraphics[scale=0.2, width=\columnwidth]{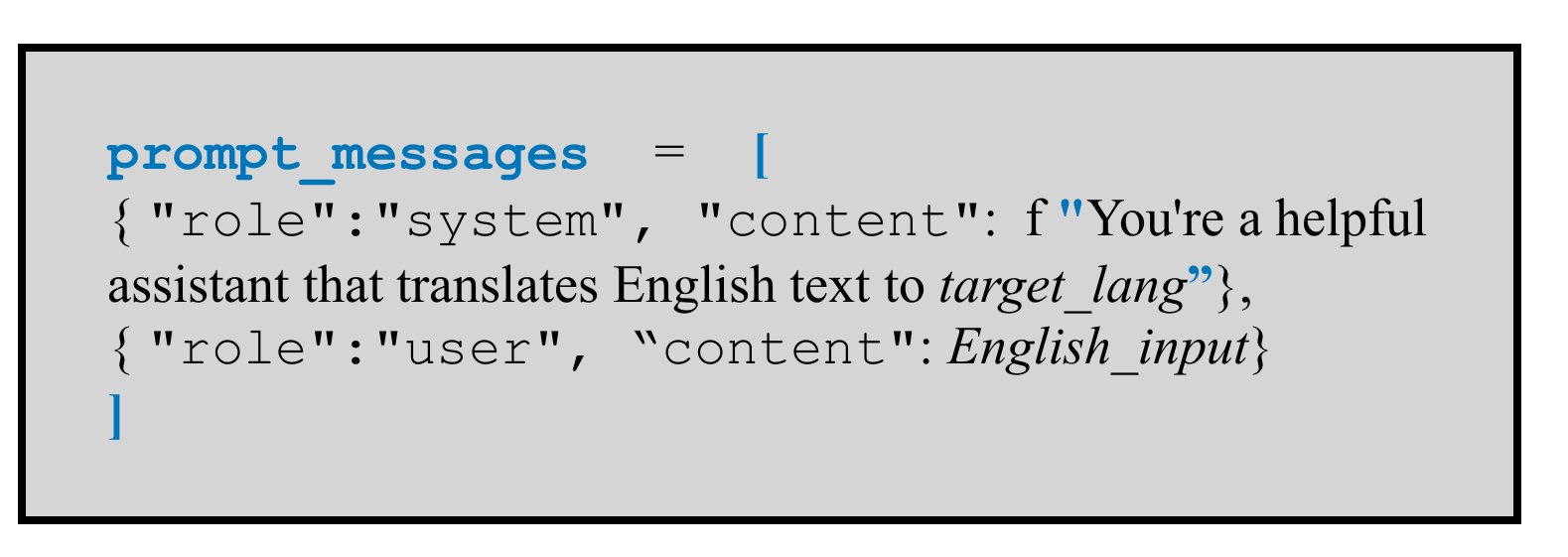}
    \caption{Prompts used for translating cultural QAs using GPT-4o.\textit{English\_input} refers to either the English question or answer to be translated into the \textit{target\_lang}, the desired target language.}
    \label{fig:QAtrans_prompts}
\end{figure}

After the curation of question-answer pairs using GPT-4o, we use the same open-source model to translate the English QA pairs into the native languages. We show the prompt that we used for this task in Fig. \textcolor{red}{A.} \ref{fig:QAtrans_prompts}.

\section{Instruction Prompt for LMM Answer Generation}
\label{sec:prompts_LMM_answer}

\begin{figure}[t]
    \centering
    \includegraphics[scale=0.2, width=\columnwidth]{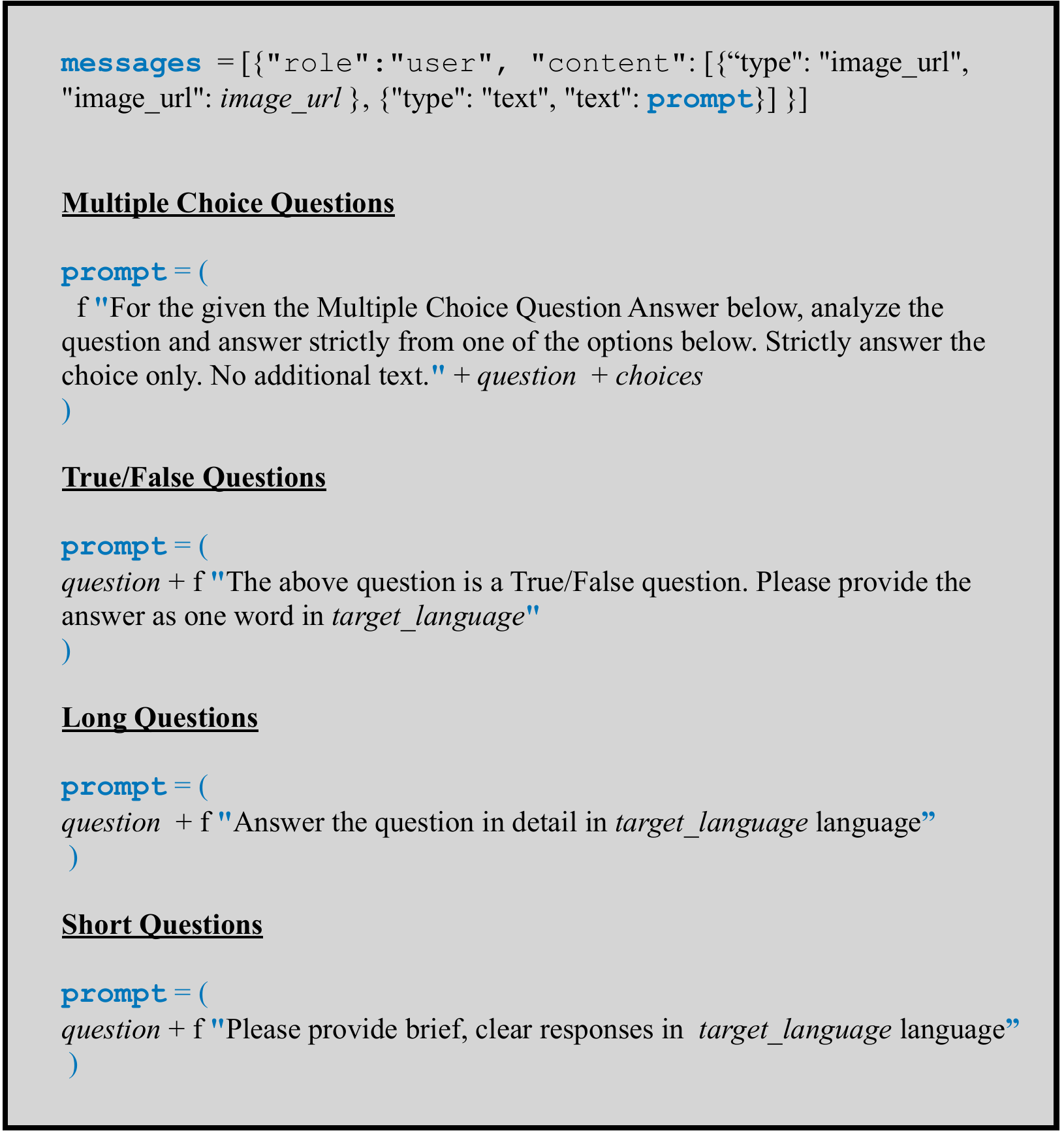}
    \caption{Prompts used to generate answers for cultural questions using multiple Large Multimodal Models evaluated in our study. Different prompts are designed for different question types. \textit{question} refers to the cultural question associated with the given image, previously generated using GPT-4o. \textit{choices} represents the four options provided in multiple choice question type, and \textit{target\_language} is the desired local language for the response.}
    \label{fig:LMMeval_prompts}
\end{figure}

We conduct a comprehensive study by evaluating various state-of-the-art Large Multimodal Models (LMMs), including both open-weight and closed-weight models, on our \ALMbench benchmark. We used different prompts for each question type as shown in Fig. \textcolor{red}{A.} \ref{fig:LMMeval_prompts}. We prompt all questions with English system instruction as highlighted by \cite{koto2024arabicmmlu}, suggesting that the use of prompts in English results in the best performance. Finally, we score each model's generated answer with the human-annotated ground truth answers through a scoring system to assess each model’s performance on the benchmark.

\section{Prompts for GPT-Scoring}
\label{sec:GPT_Scoring}

\begin{figure}[ht]
    \centering
    \includegraphics[scale=0.2, width=\columnwidth]{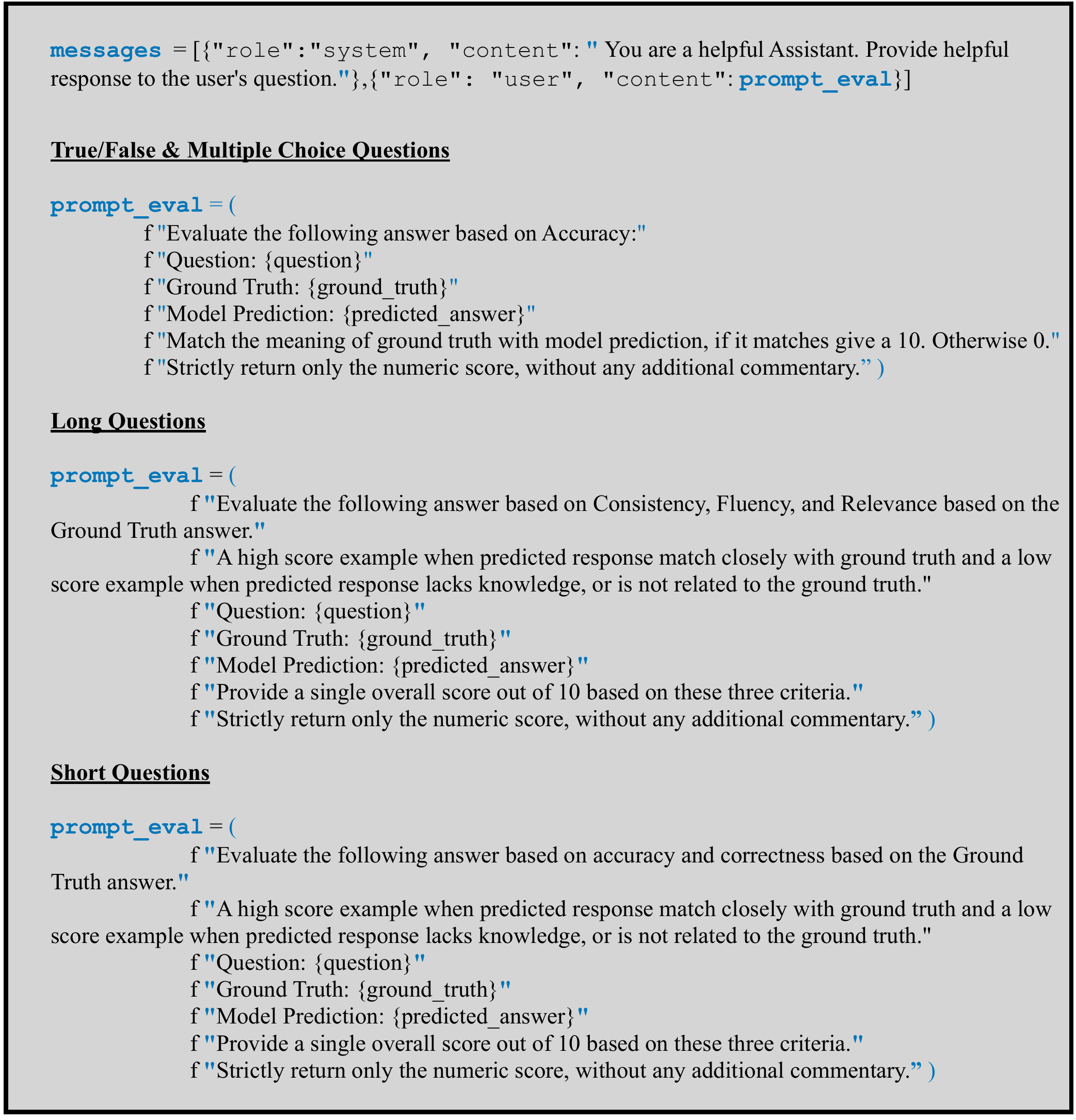}
    \caption{Prompts used to generate a score between 0 and 10, with GPT-4o acting as the evaluator to compare an LMM’s predicted answers against ground truth answers. The terms \textit{question}, \textit{ground\_truth}, and \textit{predicted\_answer} refer to the cultural question, the ground truth answer generated by GPT-4o and verified by experts, and the model’s predicted answer, respectively.}
    \label{fig:GPTscoring_prompts}
\end{figure}

In addition to using GPT-4o as a judge in scoring the predicted answers, we also do scoring with multiple other LMMs to ensure a fair evaluation. Results are reported in Table \textcolor{red}{A.} \ref{tab:decision_making_llama}.

\begin{table}[t]
\centering
\begin{tabular}{l|c|c}
\hline
Model Name & GPT-4o & Llama-3.1-8B-Instruct \\
\hline
GPT-4o          & 90.16\%   & 90.34\%   \\
Gemini-1.5 Pro  & 80.21\%   & 80.65\%   \\
GLM-4V-9B & 71.35\% & 73.89\% \\
\hline
\end{tabular}
\caption{We evaluated decision-making questions (both True/False and multiple-choice) across a sample of 20 randomly selected languages using the LLama-3.1-8B-Instruct model. This assessment aims to ensure consistency in performance on T/F and MCQs when scored using GPT-4o as a judge.}
\label{tab:decision_making_llama}
\end{table}

\section{Guidelines for Error Analysis on GPT-4o Output}
\label{sec:Error_Analysis}

Each annotator is provided with an Excel file specific to a given country-language pair, containing the following columns: Image URL, Question, Ground Truth Answer, Predicted Answer (all in the native language), and GPT-4o Score. The ‘Predicted Answer’ column records the response generated by GPT-4o when presented with the Question in the native language. The ‘GPT-4o Score’ column reflects the evaluation score assigned by GPT-4o, acting as a judge, based on the comparison between the Predicted Answer and the Ground Truth Answer.

Each annotator is required to complete two additional columns: \textbf{(1)} ‘Is the GPT Score Justified?’ with a binary response Yes/No, and \textbf{(2)} if the GPT-4o score is not justified, the ‘Reason for GPT-4o Failure’ column, where they select an appropriate reason from a predefined dropdown menu provided below.

\begin{tcolorbox}[colback=blue!5!white, colframe=blue!75!black, title=Reason for GPT-4o Failure]

\begin{itemize}
    \item \textbf{Lack of Knowledge}: The model fails to find an answer to the question.
    \item \textbf{Lack of Cultural Understanding}: The model fails to understand the cultural aspect of the answer.
    \item \textbf{Language Error}: Some words in the GPT-4o's output are wrong for the language.
    \item \textbf{Reasoning Error}: The reason it gave does not match that language.
    \item \textbf{Translation Error}: A few native words were not properly translated.
    \item \textbf{Perceptual Error}: The model fails to understand where a certain entity is (for eg, the top of the image, the bottom of the image, etc).
\end{itemize}

\end{tcolorbox}

\section{Qualitative examples with different question types}
\begin{figure*}[t!]
    \centering
    \includegraphics[width=1\textwidth]{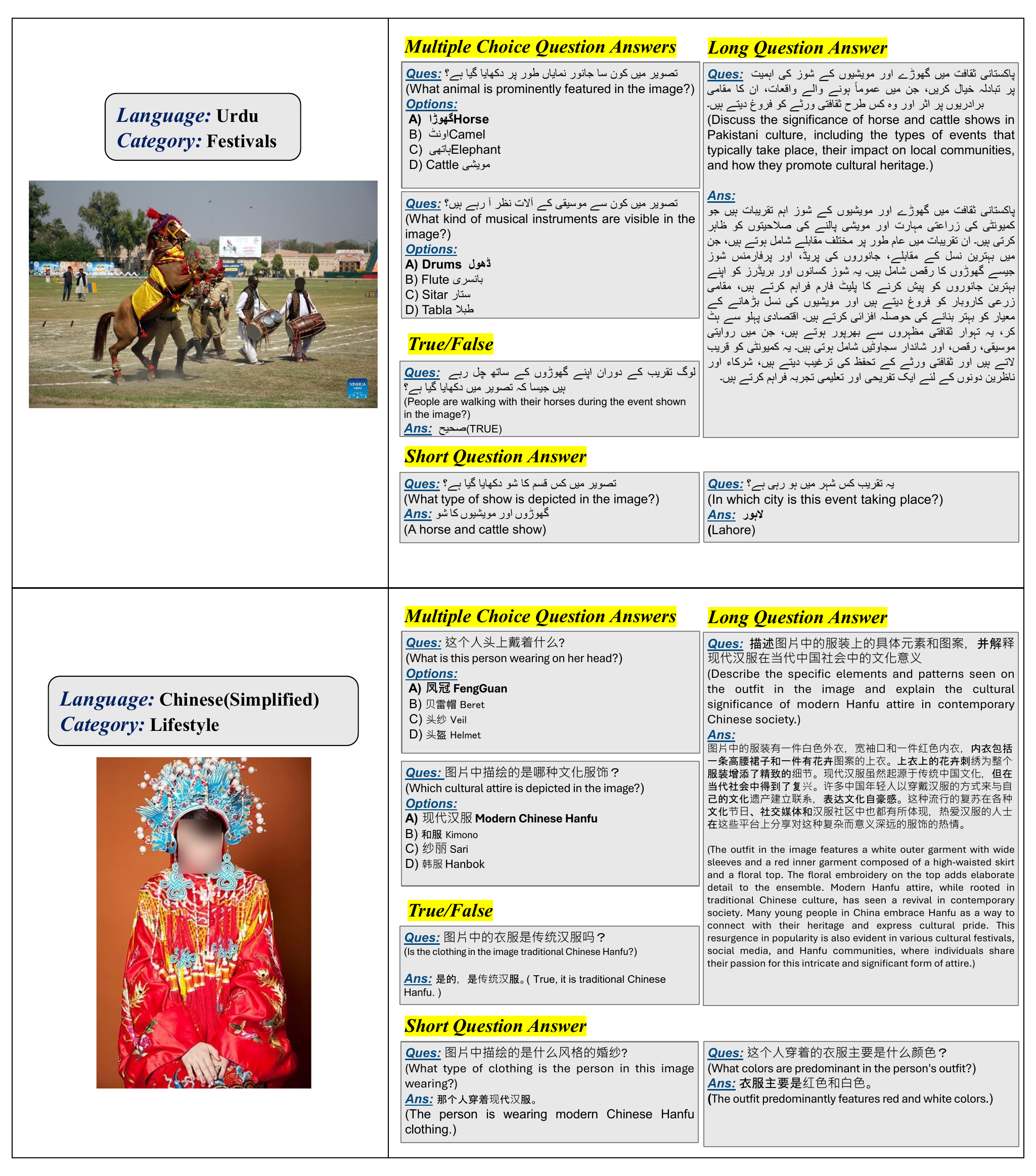}
    \caption{Qualitative examples of various question types in Urdu and Chinese Simplified Language.}
    \label{fig:qual_suppl_1}
\end{figure*}

\begin{figure*}[t!]
    \centering
    \includegraphics[width=1\textwidth]{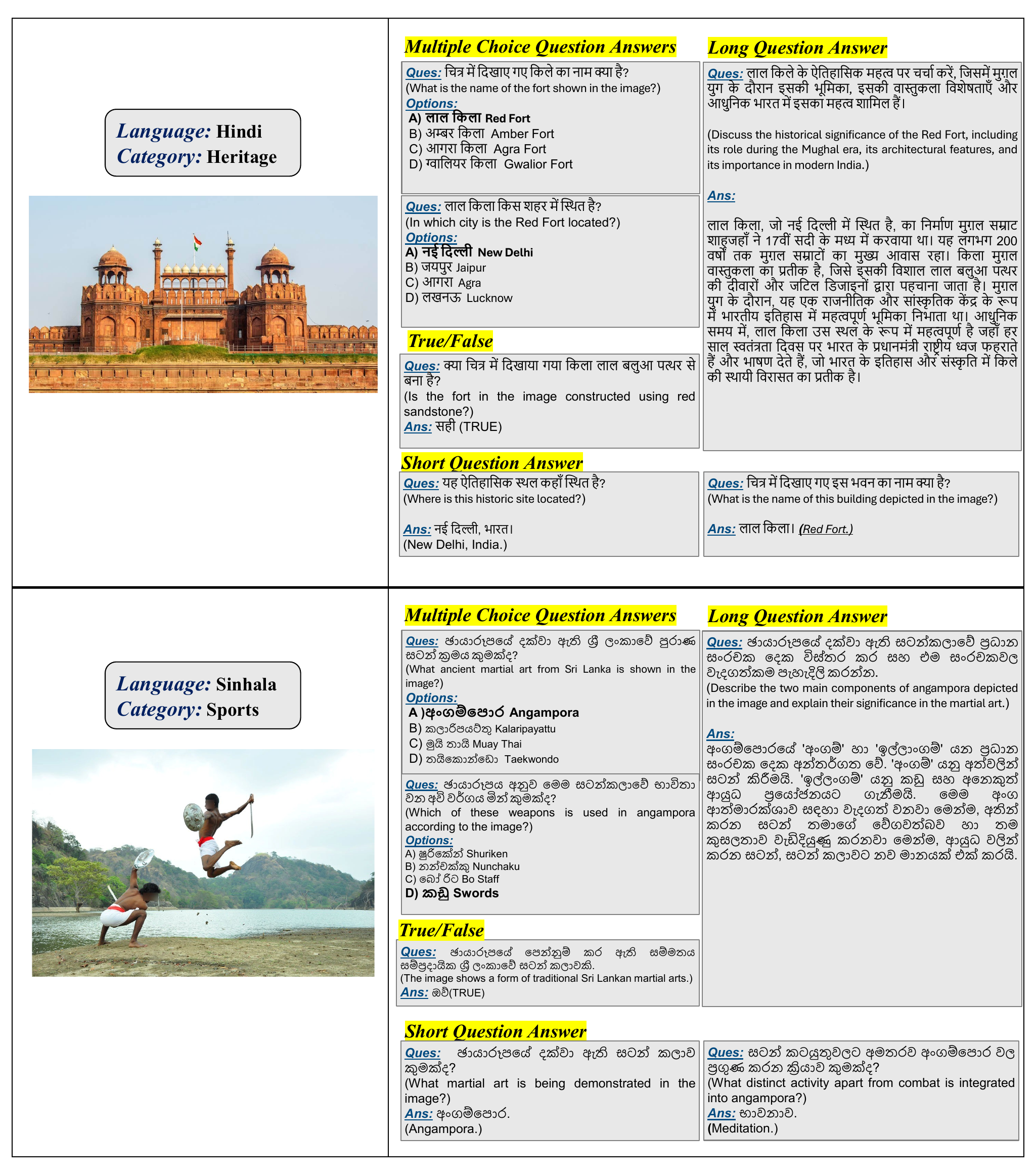}
    \caption{Qualitative examples of various question types in Hindi and Sinhala Language.}
    \label{fig:qual_suppl_2}
\end{figure*}

\begin{figure*}[t!]
    \centering
    \includegraphics[width=1\textwidth]{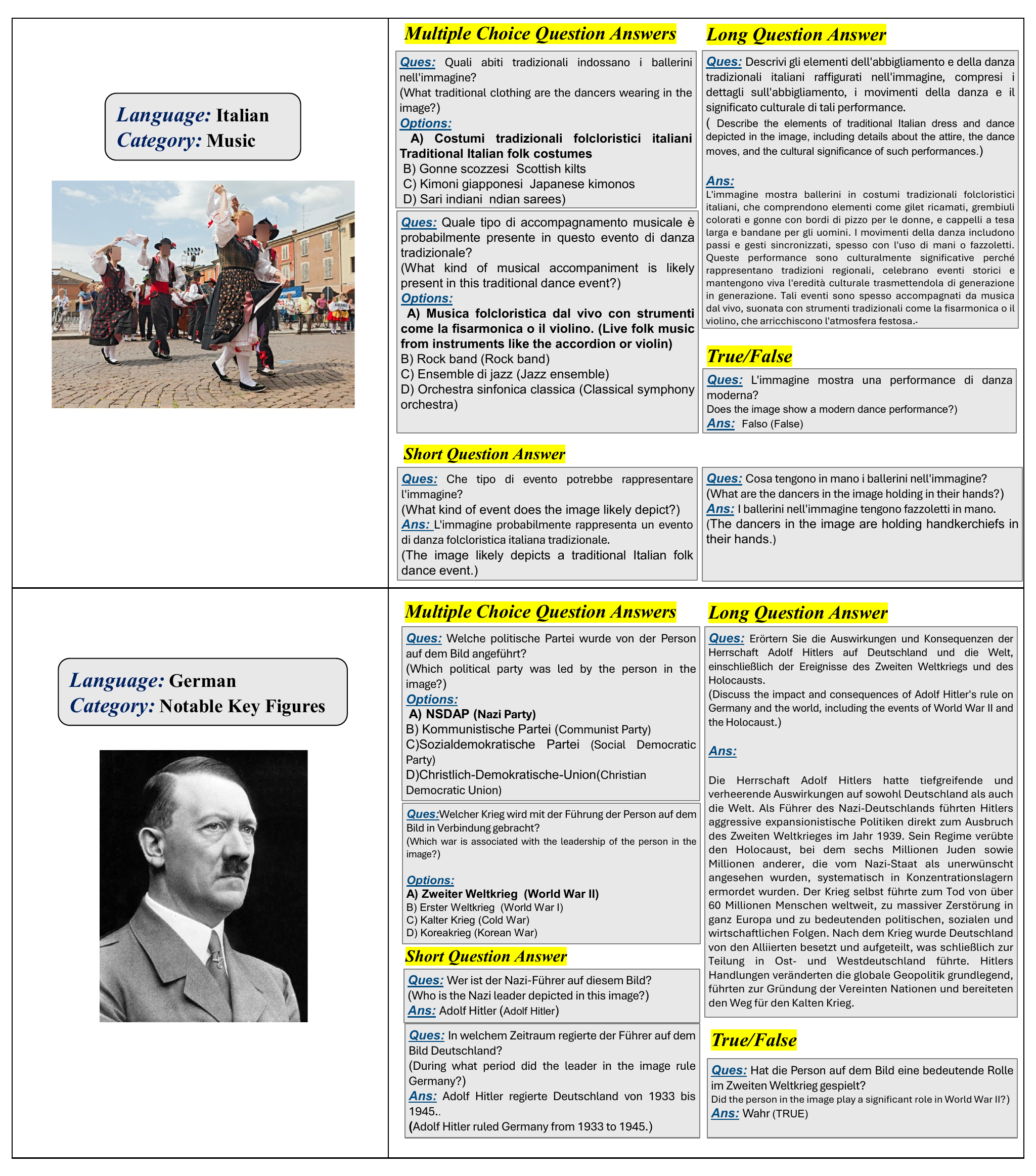}
    \caption{Qualitative examples of various question types in Italian and German Language.}
    \label{fig:qual_suppl_3}
\end{figure*}

\begin{figure*}[t!]
    \centering
    \includegraphics[width=1\textwidth]{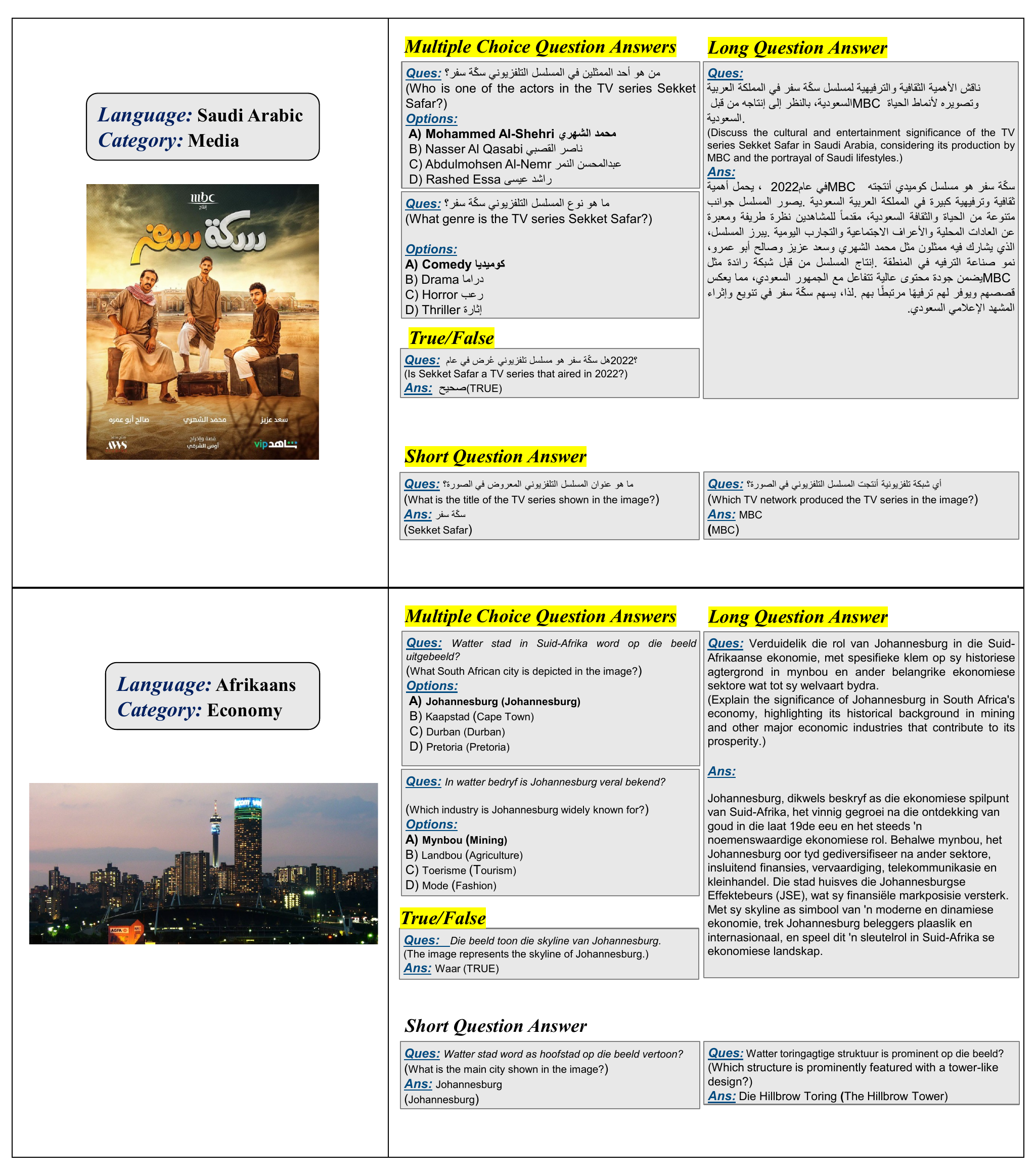}
    \caption{Qualitative examples of various question types in Saudi Arabic and Afrikaans Language.}
    \label{fig:qual_suppl_5}
\end{figure*}

\begin{figure*}[t!]
    \centering
    \includegraphics[width=0.96\textwidth]{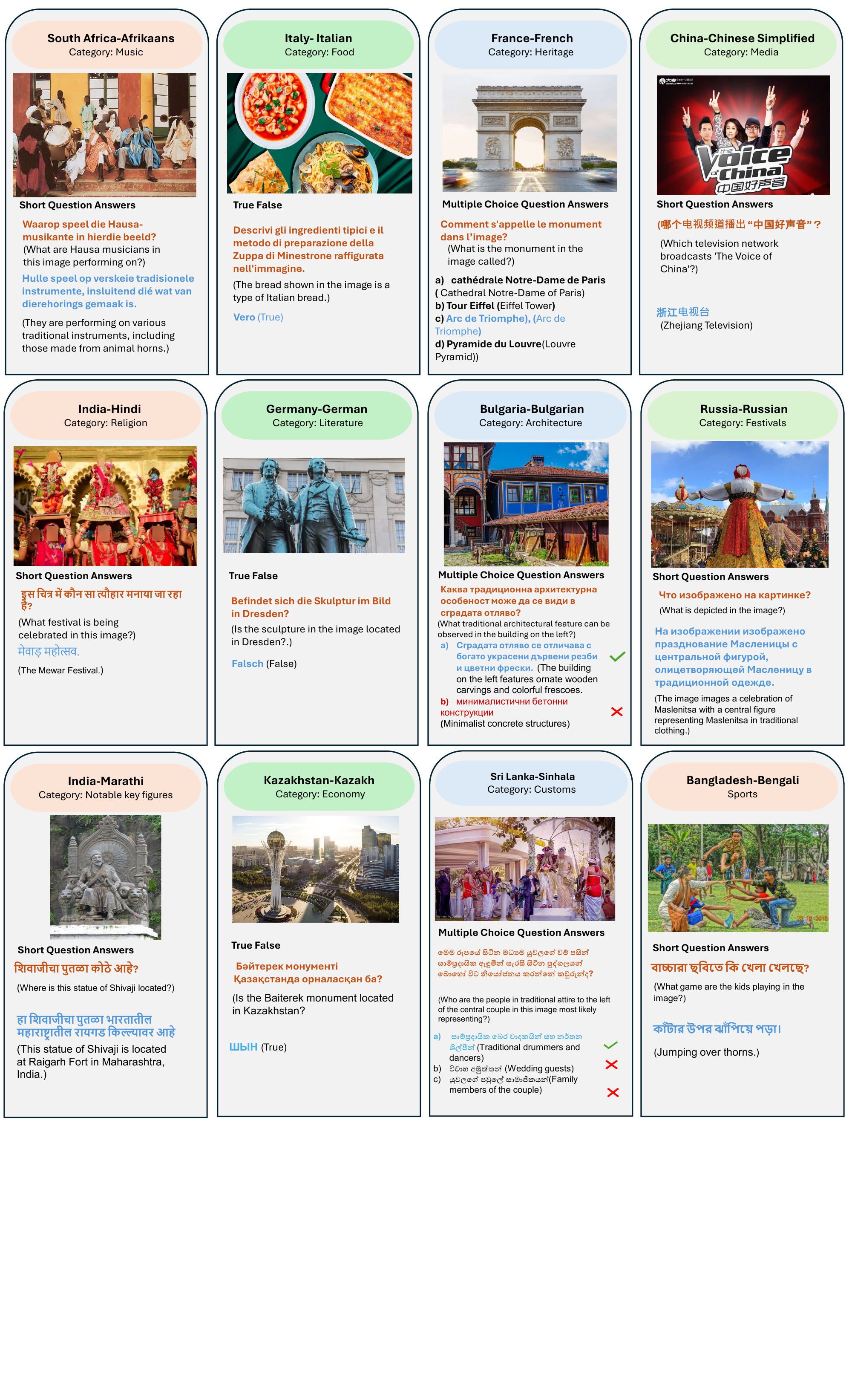}
    \caption{Some more qualitative examples of various question types from our benchmark.}
    \label{fig:suppl_categories_examples}
\end{figure*}

Next, we present some qualitative examples showing various question types for each image sample from our dataset and in Fig. \textcolor{red}{A.} \ref{fig:qual_suppl_1}, \textcolor{red}{A.} \ref{fig:qual_suppl_2}, \textcolor{red}{A.} \ref{fig:qual_suppl_3}, \textcolor{red}{A.} \ref{fig:qual_suppl_3}, \textcolor{red}{A.} \ref{fig:qual_suppl_5}, and \textcolor{red}{A.} \ref{fig:suppl_categories_examples}.

\section{Country Specific Prompts}
For the country-specific experiment as discussed in Section 4, Impact of location-aware information in prompts, Tab. 2 (main paper), we show the prompt used to conduct this experiment below. In this setting, we aim to assess how incorporating location-aware context affects vision-language model outputs, particularly in terms of potential cultural biases. To do this, we construct prompts using structured triplets comprising (language, country, cultural domain) — for example, (“Hindi,” “India,” “Food”) or (“Japanese,” “Japan,” “Religion”). This allows us to systematically evaluate whether models respond differently to culturally grounded prompts, and whether location-specific context influences model behavior or introduces/reduces bias.

\begin{tcolorbox}[colback=blue!5!white, colframe=blue!75!black, title=Location Aware Prompts]

prompt = ques + “Provide brief, clear responses in ‘lang’ language. The image represents the ‘cultural category’ in ‘country’.”

\end{tcolorbox}

\clearpage

\begin{figure*}[t!]
    \centering
    \includegraphics[width=1\textwidth]{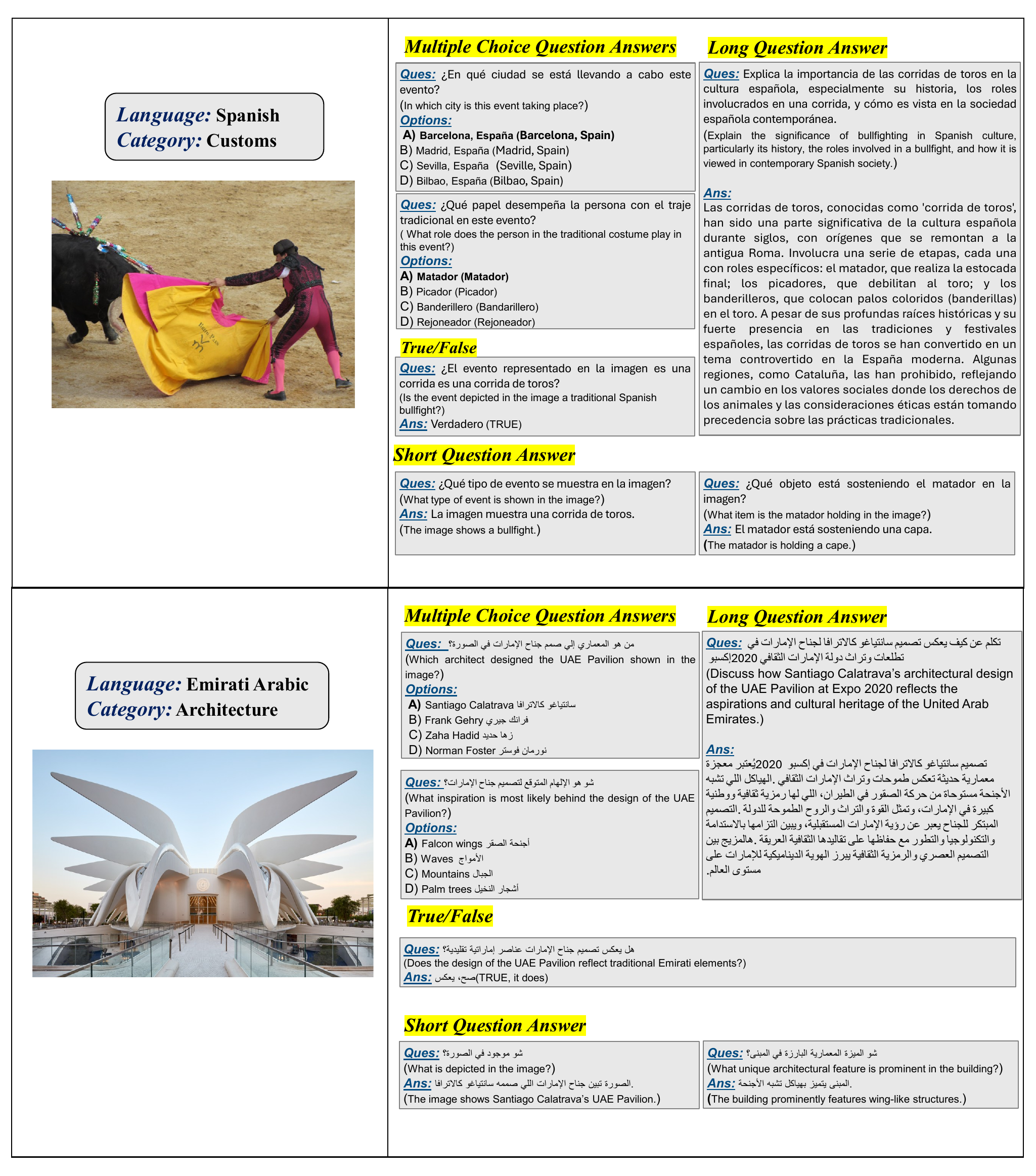}
    \caption{Qualitative examples of various question types in Spanish and Emirati Arabic Language.}
    \label{fig:qual_suppl_4}
\end{figure*}

\begin{table*}
    \centering
    \setlength{\tabcolsep}{8pt}
        \resizebox{\textwidth}{!}{%
        \begin{tabular}{llllllc}
            \toprule
             \textbf{S\#No} & \textbf{Language} & \textbf{Country} & \textbf{Script} & \textbf{Family} & \textbf{Subgrouping} & \textbf{Specification} \\

            \midrule
            1 & Afrikaans & South Africa & Latin & Indo-European & Germanic & High \\
            2 & Albanian & Albania & Latin & Indo-European & Albanian & High \\
            3 & Amharic & Ethiopia & Ge'ez & Afro-Asiatic & Semitic & Low \\
            4 & Armenian & Armenia & Armenian & Indo-European & Armenic & Low \\
            5 & Assamese & India & Bengali & Indo-European & Indo-Aryan & Low \\
            6 & Azerbaijani & Azerbaijan & Latin & Turkic & Common Turkic & Low \\
            7 & Basque & Spain & Latin & Isolate & – & High \\
            8 & Belarusian & Belarus & Cyrillic & Indo-European & Balto-Slavic & Low \\
            9 & Bengali & Bangladesh & Bengali & Indo-European & Indo-Aryan & High \\
            10 & Bhojpuri & India & Devanagari & Indo-European & Indo-Aryan & Low \\
            11 & Bosnian & Bosnia & Latin & Indo-European & Balto-Slavic & High \\
            12 & Bulgarian & Bulgaria & Cyrillic & Indo-European & Balto-Slavic & High \\
            13 & Catalan & Spain & Latin & Indo-European & Italic & High \\
            14 & Cebuano & Philippines & Latin & Austronesian & Malayo-Polynesian & Low \\
            15 & Chinese Simplified & China & Chinese & Sino-Tibetan & Sinitic & High \\
            16 & Chinese Traditional & Hong Kong & Chinese & Sino-Tibetan & Sinitic & High \\
            17 & Croatian & Croatia & Latin & Indo-European & Balto-Slavic & High \\
            18 & Czech & Czech Republic & Latin & Indo-European & Balto-Slavic & High \\
            19 & Danish & Denmark & Latin & Indo-European & Germanic & High \\
            20 & Dutch & Netherlands & Latin & Indo-European & Germanic & High \\
            21 & Egyptian Arabic & Egypt & Arabic & Afro-Asiatic & Semitic & Low \\
            22 & Emirati Arabic & United Arab Emirates & Arabic & Afro-Asiatic & Semitic & High \\
            23 & English & United Kingdom & Latin & Indo-European & Germanic & High \\
            24 & Estonian & Estonia & Latin & Uralic & Finnic & High \\
            25 & Filipino & Philippines & Latin & Austronesian & Malayo-Polynesian & Low \\
            26 & Finish & Finland & Latin & Uralic & Finnic & High \\
            27 & French & France & Latin & Indo-European & Italic & High \\
            28 & Galician & Spain & Latin & Indo-European & Italic & Low \\
            29 & Georgian & Georgia & Georgian & Kartvelian & Kartvelian & Low \\
            30 & German & Germany & Latin & Indo-European & Germanic & High \\
            31 & Greek & Greece & Greek & Indo-European & Graeco-Phrygian & High \\
            32 & Gujarati & India & Gujarati & Indo-European & Indo-Aryan & Low \\
            33 & Hausa & Nigeria & Latin & Afro-Asiatic & Chadic & Low \\
            34 & Hawaiian & United States & Latin & Austronesian & Malayo-Polynesian & Low \\
            35 & Hebrew & Israel & Hebrew & Afro-Asiatic & Semitic & High \\
            36 & Hindi & India & Devanagari & Indo-European & Indo-Aryan & High \\
            37 & Hungarian & Hungary & Latin & Uralic & – & High \\
            38 & Icelandic & Iceland & Latin & Indo-European & Germanic & High \\
            39 & Igbo & Nigeria & Latin & Atlantic-Congo & Benue-Congo & Low \\
            40 & Indonesian & Indonesia & Latin & Austronesian & Malayo-Polynesian & High \\
            41 & Irish & Ireland & Latin & Indo-European & Celtic & Low \\
            42 & Italian & Italy & Latin & Indo-European & Italic & High \\
            43 & Japanese & Japan & Kanji/Kana & Japonic & Japanese Ryukyuan & High \\
            44 & Javanese & Indonesia & Latin & Austronesian & Malayo-Polynesian & Low \\
            45 & Kannada & India & Kannada & Dravidian & South Dravidian & Low \\
            46 & Kazakh & Kazakhstan & Cyrillic & Turkic & Common Turkic & High \\
            47 & Kinyarwanda & Rwanda  & Latin & Atlantic-Congo & Benue-Congo & Low \\
            48 & Korean & South Korea & Hangul & Koreanic & Korean & High \\
            49 & Kurdish & Turkey & Arabic & Indo-European & Iranian & Low \\
            50 & Kyrgyz & Kyrgyzstan & Cyrillic & Turkic & Common Turkic & Low \\
            \bottomrule
        \end{tabular}
        }
\end{table*}
\newpage

\begin{table*}
    \centering
    \setlength{\tabcolsep}{8pt}
        \resizebox{\textwidth}{!}{%
        \begin{tabular}{llllllc}
            \toprule

            \textbf{S\#No} & \textbf{Language} & \textbf{Country} & \textbf{Script} & \textbf{Family} & \textbf{Subgrouping} & \textbf{Specification} \\

            \midrule
            51 & Lao & Thailand & Lao & Tai-Kadai & Kra-Dai & Low \\
            52 & Latin & Vatican City & Latin & Indo-European & Italic & Low \\
            53 & Latvian & Latvia & Latin & Indo-European & Balto-Slavic & High \\
            54 & Lithuanian & Lithuania & Latin & Indo-European & Balto-Slavic & High \\
            55 & Luxembourgish & Luxembourg & Latin & Indo-European & Germanic & Low \\
            56 & Macedonian & North Macedonia & Cyrillic & Indo-European & Balto-Slavic & High \\
            57 & Malagasy & Madagascar
             & Latin & Austronesian & Malayo-Polynesian & Low \\
            58 & Malay &  Malaysia & Latin & Austronesian & Malayo-Polynesian & High \\
            59 & Malayalam & India & Malayalam & Dravidian & South Dravidian & Low \\
            60 & Maltese & Malta
             & Latin & Afro-Asiatic & Semitic & High \\
            61 & Marathi & India & Devanagari & Indo-European & Indo-Aryan & Low \\
            62 & Mongolian & Mongolia & Cyrillic & Mongolic-Khitan & Mongolic & Low \\
            63 & Myanmar (Burmese) & Myanmar & Myanmar & Sino-Tibetan & Burmo-Qiangic & Low \\
            64 & Nepali & Nepal & Devanagari & Indo-European & Indo-Aryan & Low \\
            65 & Norwegian & Norway & Latin & Indo-European & Germanic & Low \\
            66 & Odia (Oriya) & India & Oriya & Indo-European & Indo-Aryan & Low \\
            67 & Pashto & Pakistan & Arabic & Indo-European & Iranian & Low \\
            68 & Persian & Iran & Arabic & Indo-European & Iranian & High \\
            69 & Polish & Poland & Latin & Indo-European & Balto-Slavic & High \\
            70 & Portuguese & Portugal & Latin & Indo-European & Italic & High \\
            71 & Punjabi & Pakistan & Gurmukhi & Indo-European & Indo-Aryan & Low \\
            72 & Romanian & Romania & Latin & Indo-European & Italic & High \\
            73 & Russian & Russia & Cyrillic & Indo-European & Balto-Slavic & High \\
            74 & Sanskrit & India & Devanagari & Indo-European & Indo-Aryan & Low \\
            75 & Saudi Arabic & Saudi Arabia & Arabic & Afro-Asiatic & Semitic & High \\
            76 & Scots Gaelic & Scotland & Latin & Indo-European & Celtic & Low \\
            77 & Serbian & Serbia & Cyrillic & Indo-European & Balto-Slavic & Low \\
            78 & Shona & Zimbabwe & Latin & Atlantic-Congo & Benue-Congo & Low \\
            79 & Sindhi & Pakistan & Arabic & Indo-European & Indo-Aryan & Low \\
            80 & Sinhala & Sri Lanka & Sinhala & Indo-European & Indo-Aryan & Low \\
            81 & Slovak & Slovakia & Latin & Indo-European & Balto-Slavic & High \\
            82 & Slovenian & Slovenia & Latin & Indo-European & Balto-Slavic & High \\
            83 & Somali & Somalia & Latin & Afro-Asiatic & Cushitic & Low \\
            84 & Spanish & Spain & Latin & Indo-European & Italic & High \\
            85 & Sundanese & Indonesia & Latin & Austronesian & Malayo-Polynesian & Low \\
            86 & Swahili & Tanzania & Latin & Atlantic-Congo & Benue-Congo & High \\
            87 & Swedish & Sweden & Latin & Indo-European & Germanic & High \\
            88 & Tajik & Tajikistan & Cyrillic & Indo-European & Iranian & Low \\
            89 & Tamil & India & Tamil & Dravidian & South Dravidian & Low \\
            90 & Telugu & India & Telugu & Dravidian & South Dravidian & Low \\
            91 & Thai & Thailand & Thai & Tai-Kadai & Kam-Tai & High \\
            92 & Turkish & Turkey & Latin & Turkic & Common Turkic & High \\
            93 & Ukrainian & Ukraine & Cyrillic & Indo-European & Balto-Slavic & High \\
            94 & Urdu & Pakistan & Arabic & Indo-European & Indo-Aryan & Low \\
            95 & Uyghur & China & Arabic & Turkic & Common Turkic & Low \\
            96 & Uzbek & Uzbekistan & Latin & Turkic & Common Turkic & High \\
            97 & Vietnamese & Vietnam & Latin & Austroasiatic & Vietic & High \\
            98 & Welsh & United Kingdom & Latin & Indo-European & Celtic & Low \\
            99 & Yiddish & Israel & Hebrew & Indo-European & Germanic & Low \\
            100 & Yoruba & Nigeria & Latin & Atlantic-Congo & Benue-Congo & Low \\

            \bottomrule
            \hline
        \end{tabular}
        }
    \caption{A comprehensive list of 100 languages, their associated country, language scripts, families, subgrouping, and the resource specification.}
    \label{tab:100_languages}
\end{table*}

\begin{table*}[ht]
\centering
\begin{tabular}{|c|l|l|l|}
\hline
\textbf{No.} & \textbf{Author Name} & \textbf{Affiliation} & \textbf{Email} \\
\hline
1 & Henok Biadglign Ademtew & Ethiopian Artificial Intelligence Institute & henokb2124@gmail.com \\
2 & Yahya Hmaiti & University of Central Florida & yohan.hmaiti@ucf.edu \\
3 & Amandeep Kumar & Johns Hopkins University & kumar.amandeep015@gmail.com \\
4 & Kartik Kuckreja & MBZUAI & kartik.kuckreja@mbzuai.ac.ae \\
5 & Mykola Maslych & University of Central Florida & mykola.maslych@ucf.edu \\
6 & Wafa Al Ghallabi & MBZUAI & wafa.alghallabi@mbzuai.ac.ae \\
7 & Mihail Mihaylov & MBZUAI & mihail.mihaylov@mbzuai.ac.ae \\
8 & Chao Qin & MBZUAI & 1746625542@qq.com \\
9 & Abdelrahman M Shaker & MBZUAI & abdelrahman.youssief@mbzuai.ac.ae\\
10 & Mike Zhang & Aalborg University & jjz@cs.aau.dk \\
11 & Mahardika Krisna Ihsani & MBZUAI & mahardika.ihsani@mbzuai.ac.ae \\
12 & Amiel Esplana & University of the West of England, Bristol & amiel2.esplana@live.uwe.ac.uk \\
13 & Monil Gokani & IIT, Hyderabad & monilgokani08@gmail.com \\
14 & Shachar Mirkin & Alpinference & shacharmirkin@gmail.com \\
15 & Harsh Singh & MBZUAI & harsh.singh@mbzuai.ac.ae \\
16 & Ashay Srivastava & University of Maryland, College Park & ashays06@umd.edu \\
17 & Endre Hamerlik & HUN-REN Institute for CS and Control & hamerlik@sztaki.hu \\
18 & Fathinah Asma Izzati & MBZUAI & fathinah.izzati@mbzuai.ac.ae \\
19 & Fadillah Adamsyah Maani & MBZUAI & fadillah.maani@mbzuai.ac.ae \\
20 & Sebastian Cavada & MBZUAI & sebastian.cavada@mbzuai.ac.ae \\
21 & Jenny Chim & Queen Mary University of London, UK & jennychim@gmail.com \\
22 & Rohit Gupta & University of Central Florida & rohitgupta.hpf@gmail.com \\
23 & Sanjay Manjunath & MBZUAI & sanjay.manjunath@mbzuai.ac.ae \\
24 & Kamila Zhumakhanova & MBZUAI & kamila.zhumakhanova@mbzuai.ac.ae \\
25 & Feno Heriniaina Rabevohitra & Chongqing University & fenoheriniaina@gmail.com \\
26 & Azril Amirudin & University of the People & azrilamirudin@my.uopeople.edu \\
27 & Muhammad Ridzuan & MBZUAI & 20020084@mbzuai.ac.ae \\
28 & Daniya Kareem & MBZUAI & daniya.kareem@mbzuai.ac.ae \\
29 & Ketan More & MBZUAI & ketan.more@mbzuai.ac.ae \\
30 & Kunyang Li & University of Central Florida & kunyang.li@ucf.edu \\
31 & Pramesh Shakya & University of Central Florida & pramesh.shakya@ucf.edu \\
32 & Muhammad Saad & MBZUAI & muhammad.saad@mbzuai.ac.ae \\
33 & Amirpouya Ghasemaghaei & University of Central Florida & aghaei.ap@ucf.edu \\
34 & Amirbek Djanibekov & MBZUAI & amirbek.djanibekov@mbzuai.ac.ae \\
35 & Dilshod Azizov & MBZUAI & dilshod.azizov@mbzuai.ac.ae \\
36 & Branislava Jankovic & MBZUAI & branislava.jankovic@mbzuai.ac.ae \\
37 & Naman Bhatia & Indian Institute of Technology, Hyderabad & naman.219311175@muj.manipal.edu \\
38 & Alvaro Cabrera & MBZUAI & alvaro.berobide@mbzuai.ac.ae \\
39 & Johan Obando-Ceron & University of Montreal, Mila & jobando0730@gmail.com \\
40 & Olympiah Otieno &  JKUAT & anikaolympiah@gmail.com \\
41 & Fabian Farestam & ETHZ - ETH Zurich & ffarestam@student.ethz.ch \\
42 & Muztoba Rabbani & University of California, Merced & mrabbani@ucmerced.edu \\
43 & Sanoojan Baliah & MBZUAI & sanoojan.baliah@mbzuai.ac.ae \\
44 & Santosh Sanjeev & MBZUAI & santosh.sanjeev@mbzuai.ac.ae \\
45 & Abduragim Shtanchaev & MBZUAI & abduragim.shtanchaev@mbzuai.ac.ae \\
46 & Maheen Fatima & Air University & 231659@students.au.edu.pk \\
47 & Thao Nguyen & MBZUAI & thao.nguyen@mbzuai.ac.ae \\
48 & Amrin Kareem & MBZUAI & amrin.kareem@mbzuai.ac.ae \\
49 & Toluwani Aremu & MBZUAI & toluwani.aremu@mbzuai.ac.ae \\
50 & Nathan Xavier & Universidade Federal de Minas Gerais & nathanxavier@ufmg.br \\
51 & Amit Bhatkal & MBZUAI & amitbhatkal12@gmail.com \\
52 & Hawau Olamide Toyin & MBZUAI & hawau.toyin@mbzuai.ac.ae\\
\hline
\end{tabular}
\caption{List of affiliations for all the volunteer co-authors who contributed to construct ALM-Bench.}
\end{table*}




\end{document}